\documentclass{fastino}

\usepackage{amsmath}
\usepackage{amssymb}
\usepackage{listings}
\usepackage[table]{xcolor}
\usepackage{appendix}
\usepackage{float}
\usepackage{multirow}
\usepackage{wrapfig}
\usepackage{tabularx}
\usepackage[ruled,vlined,linesnumbered]{algorithm2e}
\usepackage{graphicx}
\usepackage{booktabs}

\usepackage{geometry}
\usepackage{hyperref}
\usepackage{graphicx}
\usepackage{tikz}
\usetikzlibrary{shapes.geometric, arrows.meta, positioning, fit, backgrounds, calc}

\usepackage{tikz}
\usetikzlibrary{arrows.meta, positioning, fit, backgrounds, calc}

\definecolor{headergray}{RGB}{240,240,240}
\definecolor{sectiongray}{RGB}{248,248,248}
\definecolor{lightred}{RGB}{255,230,230}
\definecolor{lightgreen}{RGB}{220,250,230}
\definecolor{lightblue}{RGB}{220,235,250}

\title{Pioneer Agent: Continual Improvement of Small Language Models in Production}
\author[]{Dhruv Atreja, Julia White, Nikhil Nayak, Kelton Zhang, Henrijs Princis, George Hurn-Maloney, Ash Lewis, Urchade Zaratiana}
\affiliation[]{Fastino Labs}
\correspondence{\email{dhruv@fastino.ai}}
\date{April 10, 2026}

\abstract{
Small language models are attractive for production deployment due to their low cost, fast inference, and ease of specialization. However, adapting them to a specific task remains a challenging engineering loop, driven not by training itself but by surrounding decisions: data curation, failure diagnosis, regression avoidance, and iteration control. We present \emph{Pioneer Agent}, a closed-loop system that automates this lifecycle. In cold-start mode, given only a natural-language task description, the agent acquires data, constructs evaluation sets, and iteratively trains models by jointly optimizing data, hyperparameters, and learning strategy. In production mode, given a deployed model with labeled failures, it diagnoses error patterns, constructs targeted training data, and retrains under explicit regression constraints. To evaluate this setting, we introduce \emph{AdaptFT-Bench}, a benchmark of synthetic inference logs with progressively increasing noise, designed to test the full adaptation loop: diagnosis, curriculum synthesis, retraining, and verification. Across eight cold-start benchmarks spanning reasoning, math, code generation, summarization, and classification, Pioneer Agent improves over base models by 1.6--83.8 points. On AdaptFT-Bench, it improves or preserves performance in all seven scenarios, while naive retraining degrades by up to 43 points. On two production-style deployments built from public benchmark tasks, it raises intent classification from 84.9\% to 99.3\% and Entity~F1 from 0.345 to 0.810.
Beyond performance gains, the agent often discovers effective training strategies, including chain-of-thought supervision, task-specific optimization, and quality-focused data curation, purely from downstream feedback.
}
\begin{document}

\maketitle

\section{Introduction}
Frontier large language models \citep{brown2020gpt3,touvron2023llama,yang2025qwen3technicalreport,DeepSeekAI2025DeepSeekR1IR} are strong general-purpose systems \citep{bommasani2021foundation}. However, they are often too expensive to deploy and operate at scale \citep{kaplan2020scaling}, and can exhibit high latency at inference time compared to smaller models \citep{fu2025nemotronflash}. In practice, teams often want a much smaller model that is cheap to serve, fast at inference, and specialized for one concrete job such as intent classification \citep{larson2019evaluation,alekseev-etal-2025-autointent}, entity extraction \citep{zaratiana-etal-2022-global}, summarization \citep{rush-etal-2015-neural,chen2018fast}, or structured generation \citep{chen2021evaluating}. Small language models (SLMs) in the 1B--8B parameter range are attractive for exactly this reason \citep{dettmers2023qlora}.

The problem is that getting a small model to work well on a specific task is still a difficult engineering loop. This requires deciding what data to collect, which base model and training recipe to use, how to evaluate progress, which failures matter, and when to stop iterating. These decisions reflect broader challenges studied in AutoML and neural architecture optimization \citep{feurer2019automl,elsken2019neural}.

This loop extends beyond hyperparameter tuning. In real deployments, the primary challenges lie upstream of training: defining the task precisely, constructing and cleaning supervision, identifying boundary cases, separating true failures from labeling errors, and avoiding regressions while fixing known weaknesses. Improvements are often non-monotonic: a model can improve on one slice while degrading on another, and a larger dataset can underperform a smaller but higher-quality one \citep{zhou2023limaalignment, Iyer2024QualityOQ}. Even prompt-format issues can masquerade as modeling failures. For small models, these decisions frequently dominate the final outcome \citep{Tang2024LargeLM, Ngweta2025TowardsLR}.

Existing work addresses important pieces of this problem but not the entire loop. \cite{pareja2024unveiling} characterize effective supervised fine-tuning configurations for small LLMs, but applying these insights in practice still requires substantial manual iteration. AutoML systems \citep{thornton2013autoweka, feurer2015autosklearn} automate hyperparameter selection over predefined spaces. Data-centric methods \citep{northcutt2021confident, mazumder2023dataperf, wang2023letssynthesizestepstep} focus on label quality and iterative data improvement. Prompt and LM-program optimizers \citep{khattab2023dspy, opsahl-ong2024optimizing, yuksekgonul2024textgrad} tune the interface to a fixed model rather than the model weights. On the agentic side, execution-grounded search has proven effective for GPU kernel optimization \citep{chen2026avo} and scientific discovery \citep{yamada2025ai, gottweis2025towards}, while ML engineering benchmarks \citep{chan2024mlebench, nam2025mle} and autonomous fine-tuning environments \citep{li2026ft, rank2026posttrainbench} show that LLM agents can execute substantial portions of the ML workflow. These systems are reviewed in detail in Section~\ref{sec:related-work}. However, they primarily address offline optimization and do not center the production adaptation problem: starting from real inference failures, determining which are fixable by training, constructing targeted supervision, retraining, and verifying that improvements do not introduce regressions.

\begin{figure}
\centering
\includegraphics[width=.9\linewidth]{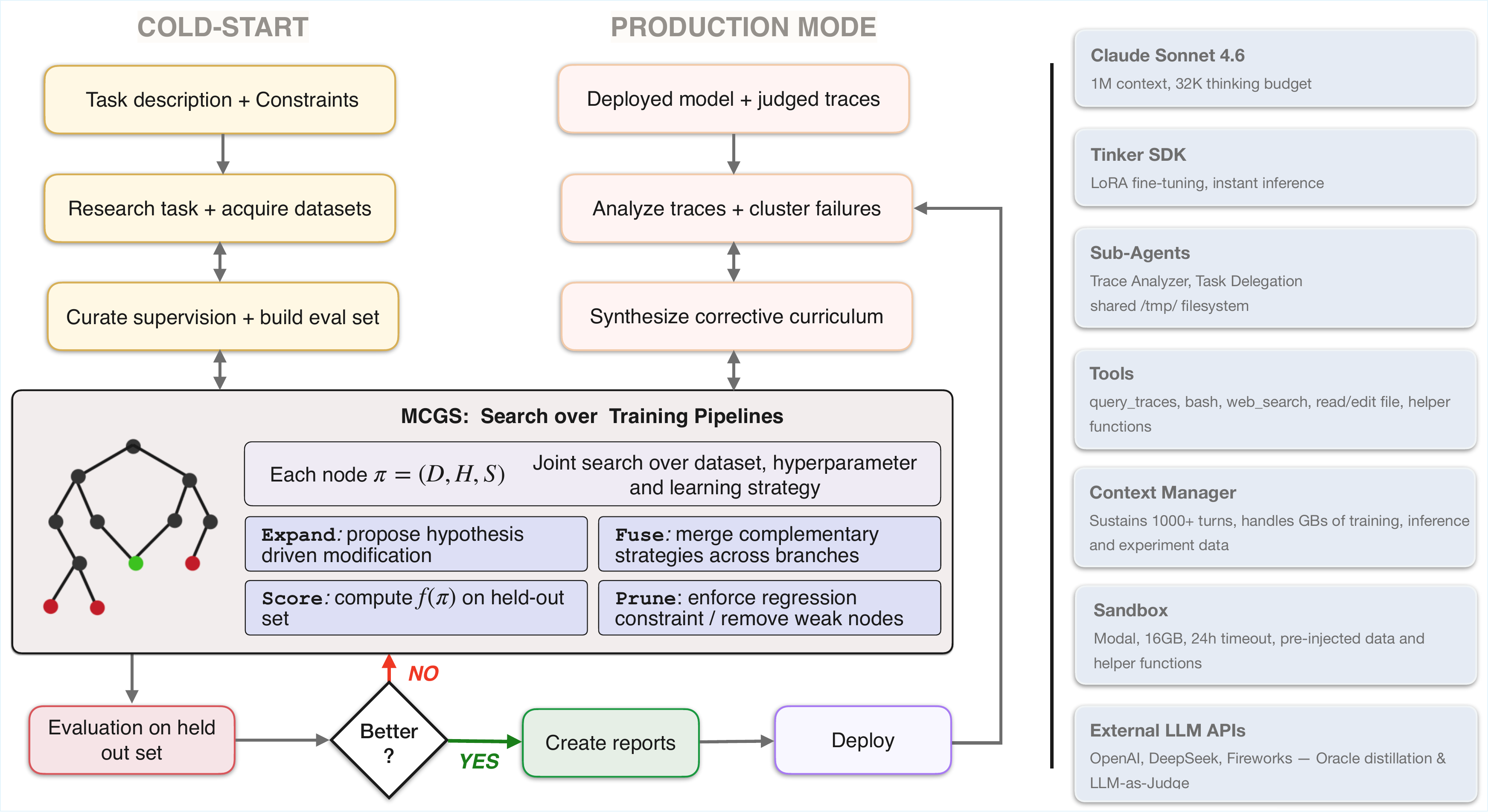}
\caption{\textbf{Pioneer Agent system architecture.} An orchestrator LLM (Claude Sonnet 4.6) drives a LangGraph state machine coordinating task analysis, data curation, training, and evaluation. Cold-start mode (left) starts from a task description; production mode (right) uses judged inference traces. Both share an agent-guided iterative search over training pipelines $\pi = (D, H, S)$, with explicit graph-structured MCGS in selected runs such as ARC-Challenge, executed in Modal sandboxes with the Tinker SDK.}
\label{fig:architecture}
\end{figure}


In this paper, we study that setting directly. We present the \textbf{Pioneer Agent}, a closed-loop system for autonomous SLM adaptation. The system operates in two modes. In \emph{cold-start mode}, the input is only a natural-language task description, and the agent builds a fine-tuned model from scratch by (1) researching the task, (2) identifying and downloading real datasets—augmenting with synthetic data when needed, (3) constructing a held-out validation set before training, (4) curating supervision with structured guidelines (e.g., hard negatives, label balancing, and context-length matching), (5) training multiple configurations in parallel, and (6) iterating against the held-out validation set. In \emph{production mode}, the input is a deployed model together with judged inference failures (from LLM-as-judge or human review). The agent repairs these failures by querying production traces, constructing a failure taxonomy, confirming weaknesses via targeted probing, synthesizing a corrective training curriculum (including corrected examples, hard negatives, and replay data), retraining the model, and evaluating against both the failure set and a regression set, and iterating until failures are resolved without degrading existing behavior. In Pioneer Agent, the search space includes more than model hyperparameters. The agent searches jointly over training data composition, hyperparameter configurations, and learning strategies. We implement this search using agent-guided iterative search. In selected runs (e.g., ARC-Challenge), this takes the form of \textbf{Monte Carlo Graph Search (MCGS)}, a graph-based extension of Monte Carlo Tree Search. Each node corresponds to a complete training attempt and its measured outcome, allowing the agent to reason over trajectories of experiments rather than isolated trials. This enables it to distinguish data problems from optimization problems and to combine useful strategies discovered across different search branches.

\begin{figure}
\centering
\includegraphics[width=0.95\linewidth]{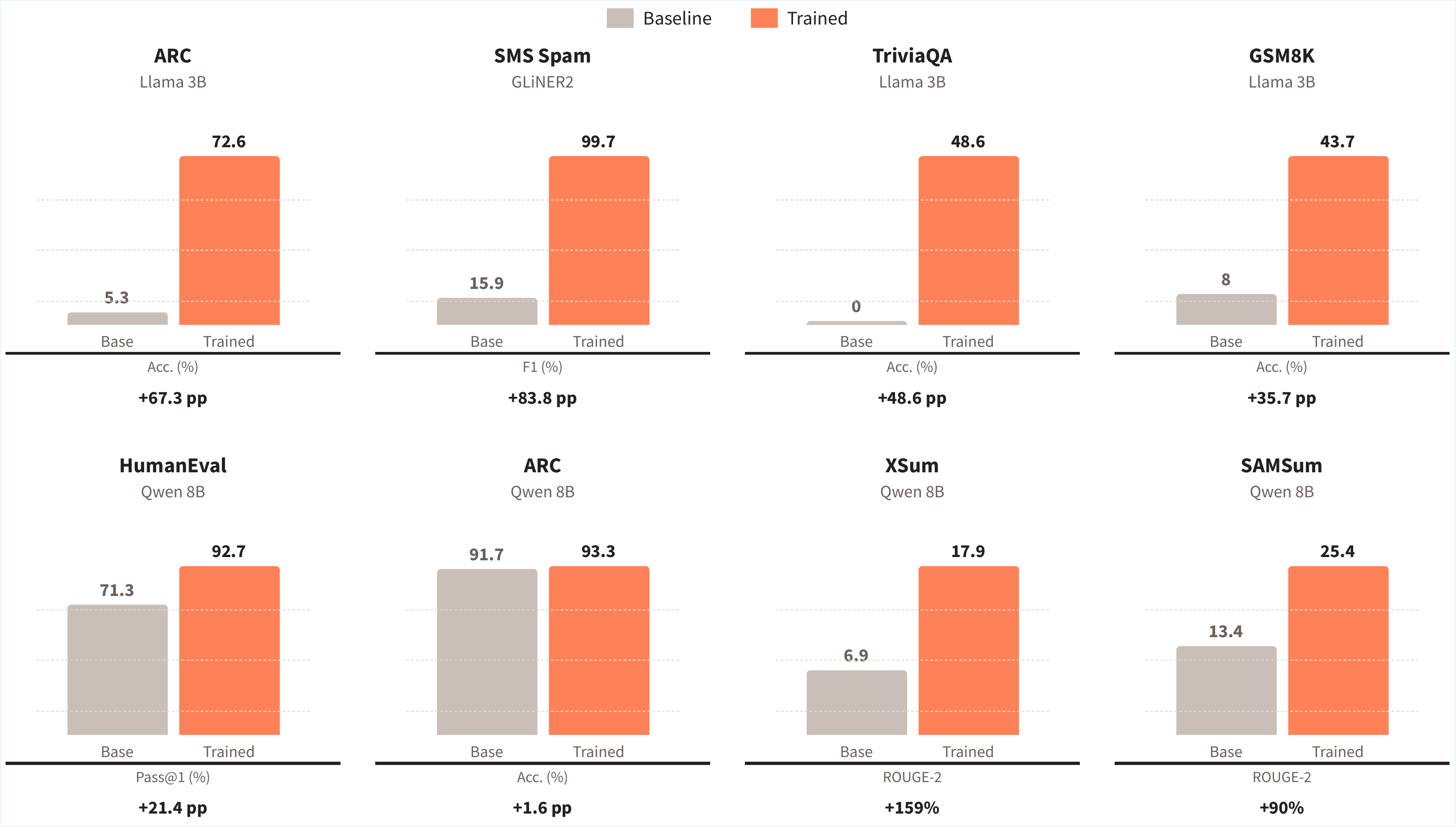}
\caption{\textbf{Cold-start performance: baseline vs.\ Pioneer Agent.}
Each pair of bars compares the base model score (grey) with the Pioneer Agent--trained score (orange) on the task-specific metric (accuracy, F1, pass@1, or ROUGE-2).
Results span three model families (Llama\,3B, Qwen\,8B, GLiNER2) and four task categories: reasoning (ARC, TriviaQA, GSM8K), classification (SMS Spam), code generation (HumanEval), and summarization (XSum, SAMSum).}
\label{fig:benchmark-results}
\vspace{-1em}
\end{figure}

The production setting changes what counts as improvement. Offline benchmark gains alone are insufficient. An effective adaptation system must operate on noisy and heterogeneous failure signals, identify whether errors arise from data issues, model limitations, or labeling noise, and satisfy regression constraints on previously correct behavior. Pioneer Agent is designed to meet these requirements. Our contributions are as follows:

\begin{enumerate}
    \item We present \textbf{Pioneer Agent}, a closed-loop system for autonomous adaptation of SLMs, supporting both cold-start task bootstrapping from natural language descriptions and production-time improvement from observed failures.

    \item We introduce a \textbf{failure-driven curriculum synthesis method} that constructs training data from production inference failures, using failure taxonomy construction, weakness confirmation via targeted probing, and structured dataset composition (corrected examples, hard negatives, and replay data).

    \item We introduce \textbf{AdaptFT-Bench}, a benchmark built from synthetic inference logs that evaluates autonomous model improvement across the full loop of failure diagnosis, curriculum synthesis, retraining, and regression verification.

    \item We demonstrate strong empirical results across eight cold-start benchmarks and multiple production scenarios, including substantial gains on math, coding, summarization, and classification tasks.

    \item We show that the agent consistently selects effective fine-tuning strategies---including chain-of-thought training for reasoning \citep{wei2022chain}, task-specific epoch sensitivity, quality-over-quantity data curation, and system prompt engineering---based on observed performance feedback, without explicit instruction to use any particular technique.
\end{enumerate}

\section{Pioneer Agent}
\label{sec:pioneer-agent}

\subsection{Architecture}

Pioneer Agent is built on a LangGraph state machine orchestrated by Claude Sonnet 4.6 \citep{anthropic2026sonnet46} with extended thinking (32K token budget) and a 1M token context window. The agent executes in Modal sandboxes -- isolated containers with 16GB memory and up to 24-hour timeouts -- with access to a suite of tools for data manipulation, model training, and evaluation.

The system supports two complementary model families, corresponding to encoder-based and decoder-based architectures, which together cover classification, extraction and generation tasks:
\begin{itemize}
    \item \textbf{GLiNER2 (encoder)}: For named entity recognition (NER) and text classification. GLiNER2 \citep{zaratiana-etal-2025-gliner2} extends GLiNER \citep{zaratiana-etal-2024-gliner} with multi-task training over NER and classification tasks. It retains the span-based architecture for NER and introduces a dedicated classification head trained jointly with the encoder. The model uses shared representations with task-specific output layers for each task. Variants include \texttt{gliner2-base-v1}, \texttt{gliner2-large-v1}, and \texttt{gliner2-multi-v1} (multilingual). Training completes in 2--5 minutes.
    \item \textbf{Qwen/Llama (decoder)}: For text generation tasks. All Qwen3-8B \citep{qwen3technicalreport} experiments use the instruction-tuned variant; all Llama-3.2-3B \citep{grattafiori2024llama3herdmodels} experiments use the base (non-instruction-tuned) variant. Training takes 10--30 minutes depending on model size.
\end{itemize}

Decoder training is performed via the Tinker SDK, which provides Low-Rank Adaptation (LoRA) \citep{hu2021loralowrankadaptationlarge} fine-tuning with instant inference on the resulting model. Encoder training (GLiNER2) supports both full fine-tuning and LoRA, as the smaller encoder architecture makes full parameter updates practical. The agent interacts with a set of pre-loaded helper functions (\texttt{tinker\_helpers.py}) for dataset management, training orchestration, and inference.

The agent uses a hierarchical architecture to manage context and parallelize work. The main agent orchestrates the full pipeline for up to 500 (production) or 1,500 (cold-start) LangGraph turns, where each turn is one LLM reasoning step together with its associated tool calls. A specialized \emph{Trace Analyzer} sub-agent handles data-heavy SQL analysis with a ${\sim}$100K output-token limit, and independent sub-agents can be spawned via \texttt{delegate\_task} for parallel work (e.g., building the next dataset while a training job runs). A proprietary \emph{Context Manager} module monitors the conversation state and ensures the agent can sustain hundreds of reasoning turns without losing track of prior decisions. When the conversation approaches the context limit, the module selectively compacts older turns while preserving key decisions, evaluation results, and dataset lineage. The internal mechanism of the Context Manager is not disclosed in this paper. The agent also maintains a persistent \texttt{data-curation.md} log---a structured markdown file recording dataset versions, composition ratios, quality-control decisions, and per-iteration evaluation results---that is written to disk and survives compaction cycles, providing durable provenance that the agent can re-read at any point.

\subsection{Search Procedure}
\label{sec:search-space}

We formalize autonomous fine-tuning as search over a structured configuration space that jointly captures data, optimization, and supervision choices. A \emph{training pipeline} is defined as a tuple $\pi = (D, H, S)$, where each component specifies a distinct but interdependent aspect of the training process:

\begin{itemize}
    \item $D$ is a \emph{dataset specification}: the composition of gold examples, hard negatives, and replay data, together with the curation constraints applied (Section~\ref{sec:data-curation}).
    \item $H$ is a \emph{hyperparameter configuration}: base model, LoRA rank, learning rate, batch size, number of epochs, and system prompt.
    \item $S$ is a \emph{learning strategy}: the supervision format (direct answer vs.\ chain-of-thought), the teacher model used for annotation (if any), and the evaluation method.
\end{itemize}

Let $\Pi$ denote the space of all valid pipelines. We define a scoring function $f\colon \Pi \to \mathbb{R}$ that maps a pipeline to its held-out validation-set performance by executing the full train-then-evaluate loop. The score is the task-appropriate metric: accuracy or F1 for classification and NER, and an LLM-as-judge score normalized to $[0,1]$ for generation tasks (summarization, code, QA). We also define a regression function $r\colon \Pi \times 2^{\mathcal{X}} \to \mathbb{Z}_{\geq 0}$ that counts the number of previously correct examples that the new model answers incorrectly on a held-out regression set $\mathcal{R} \subset \mathcal{X}$. The optimization objective is:
\begin{equation}
    \pi^* = \arg\max_{\pi \in \Pi}\; f(\pi) \quad \text{subject to} \quad r(\pi;\, \mathcal{R}) \leq \epsilon
    \label{eq:objective}
\end{equation}
where $\epsilon = 2$ in practice. We adopt an absolute count rather than a relative rate for the regression constraint: in a production setting, even a single systematic regression pattern warrants investigation, whereas a relative threshold would permit proportionally more regressions on larger test sets. This conservative choice prioritizes deployment safety over search flexibility. The regression constraint is enforced only in production mode; in cold-start mode, where no prior model exists, the unconstrained objective $\pi^* = \arg\max_{\pi \in \Pi} f(\pi)$ is used instead.

\begin{figure}
\centering
\includegraphics[width=0.96\linewidth]{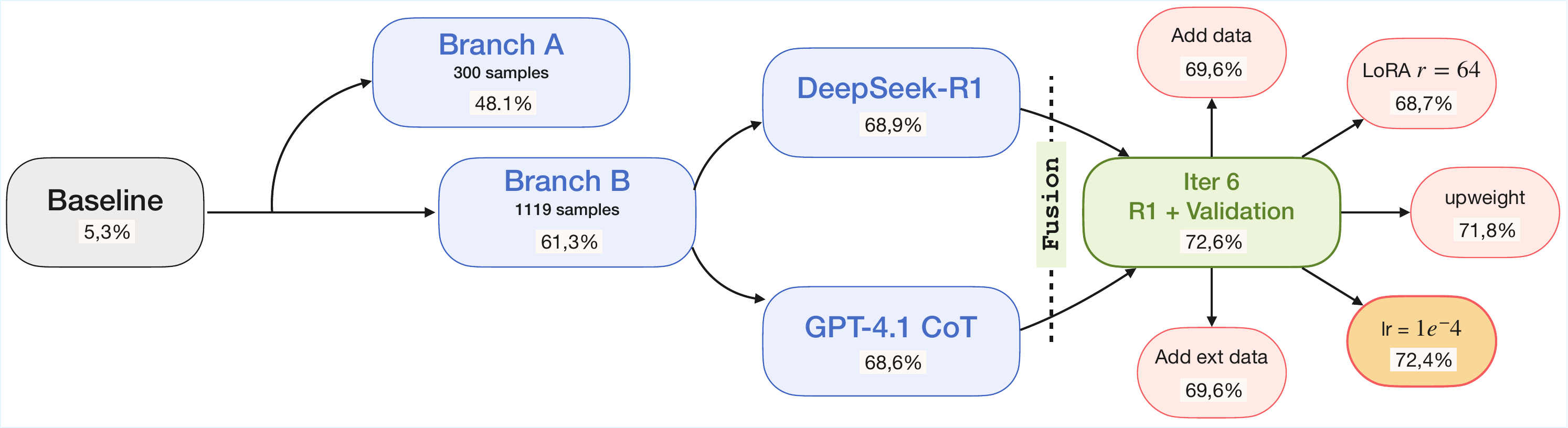}
\caption{\textbf{Monte Carlo Graph Search trajectory for ARC-Challenge (Llama 3.2-3B) over 11 iterations.} Each node represents a completed training pipeline $(\pi_i, f(\pi_i))$; edges encode the agent's causal reasoning about why a configuration succeeded or failed. The graph shows three key phases: format learning (Branch~A, 5.3\%$\to$61.3\%), chain-of-thought discovery (Branch~B, 68.6\%), and cross-branch fusion of DeepSeek-R1 reasoning traces with validation-data expansion (Iteration~6, 72.6\%). Post-fusion expansions (Iterations 7--11) all regress and are pruned, confirming convergence.}
\label{fig:mcgs-arc}
\end{figure}

\begin{wrapfigure}{r}{0.5\textwidth}
\vspace{-\intextsep}
\centering
\caption{\textbf{Pioneer Agent's two operational modes.} \emph{Cold-start mode} (top) takes a natural-language task specification and autonomously downloads data, constructs evaluation sets, synthesizes training curricula, and iterates via agent-guided search (MCGS in selected runs). \emph{Production mode} (bottom) takes a deployed model with judged inference failures, constructs a failure taxonomy, synthesizes corrective training data (gold corrections, hard negatives, replay buffer), retrains, and accepts updates only when they pass both the failure evaluation set and a regression gate on previously correct behavior.}
\includegraphics[width=0.51\textwidth]{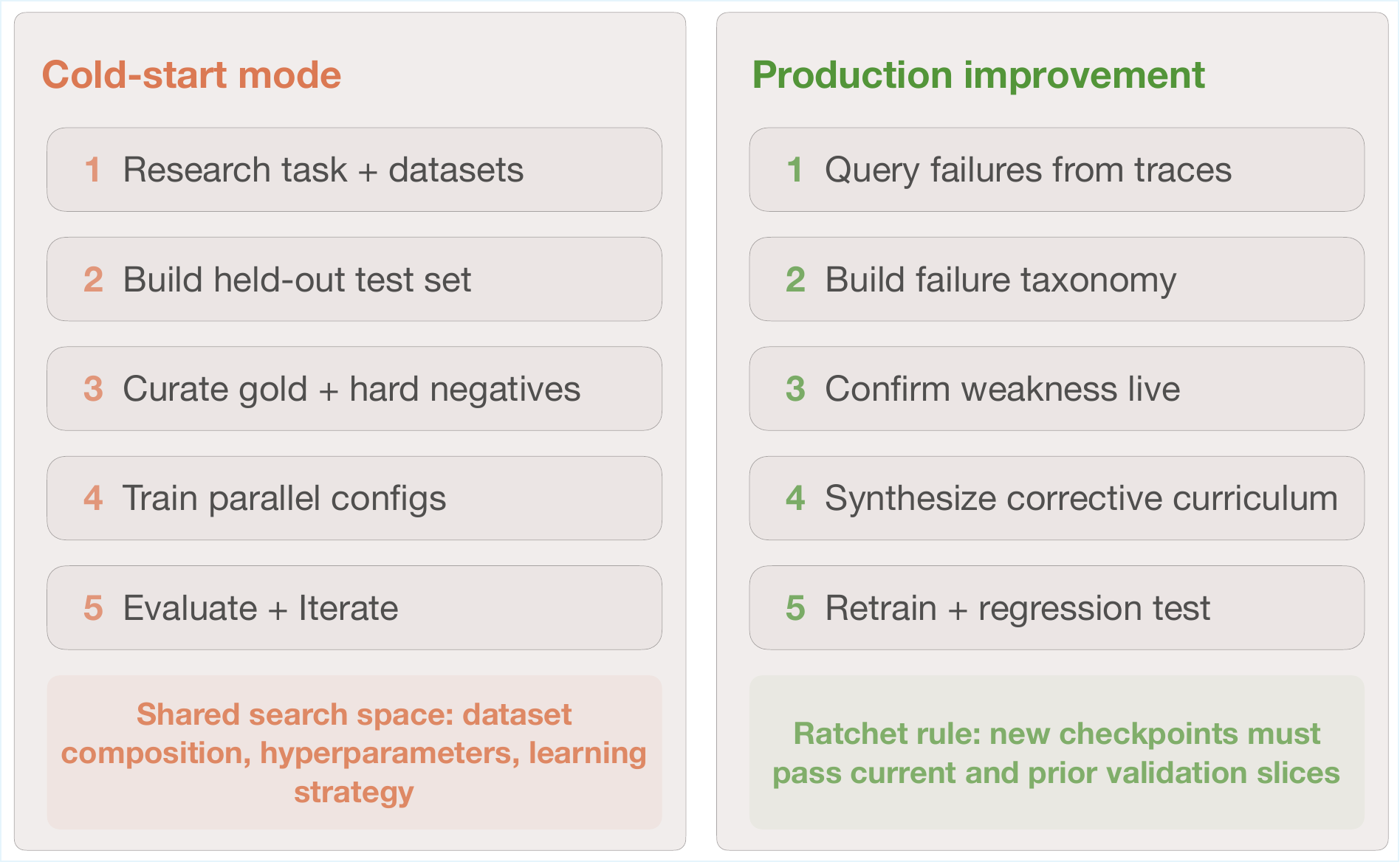}

\label{fig:operational-modes}
\vspace{-1em}
\end{wrapfigure}

The function $f$ has no closed-form expression and no useful gradient -- each evaluation requires training a model and running inference on the full validation set. The search space $\Pi$ is also heterogeneous: $D$, $H$, and $S$ interact in ways that cannot be decomposed \citep{feurer2022autosklearn20handsfreeautoml} (e.g., chain-of-thought supervision requires more epochs than direct-answer supervision; larger datasets may require lower learning rates). This rules out standard optimization over a fixed parameter grid (like \textit{grid search} \citep{hsu2003practical}  or \textit{random search} \citep{10.5555/2188385.2188395}) and motivates a search procedure that reasons about \emph{why} a configuration succeeded or failed.

The agent navigates $\Pi$ using Monte Carlo Graph Search (MCGS; \citealp{du2025automlgen, feng2026internagent}). Unlike traditional hyperparameter search, which treats each configuration independently, MCGS maintains a directed acyclic graph $\mathcal{G} = (\mathcal{V}, \mathcal{E})$ that encodes relationships between training attempts. Each node $v_i = (\pi_i,\, f(\pi_i))$ records a completed \emph{iteration}---constructing or modifying the dataset, training a model, and evaluating it on a held-out set. An edge $(v_i, v_j) \in \mathcal{E}$ indicates that $\pi_j$ was derived from $\pi_i$ via a targeted modification (e.g., data curation, hyperparameter change, or supervision shift). This structure allows the agent to reason over trajectories rather than isolated trials. By preserving lineage, it can attribute performance changes to specific interventions, reuse successful patterns, and avoid unproductive regions of the search space. The graph thus captures not only \emph{what} was tried but also \emph{why}.

This causal structure determines how $\mathcal{G}$ grows. At each step the agent selects a leaf node $v_{\mathrm{parent}}$ and proposes a child configuration via an expansion operator:
\begin{equation}
    \pi_{\mathrm{child}} = \textsc{Expand}(v_{\mathrm{parent}},\, \mathcal{G},\, \mathcal{F})
    \label{eq:expand}
\end{equation}
where $\mathcal{F}$ is the failure set or task specification. $\textsc{Expand}$ is implemented by the orchestrator LLM (aka the \textsc{Agent}): it inspects the parent trajectory, diagnoses failure modes, and proposes a hypothesis-driven modification---for example, reducing epochs in response to overfitting or introducing hard negatives to resolve systematic label confusion. Because each new experiment is a reasoned response to observed failures rather than a stochastic perturbation, the search is both more efficient and more interpretable than random alternatives.

Every child node is scored by executing the full pipeline: $f(\pi_{\mathrm{child}})$ trains the model and evaluates it on the held-out validation set. While \emph{surrogate models} (for example, predictors that estimate final validation accuracy from early training loss or truncated runs) could reduce cost, they introduce bias that is difficult to quantify in the presence of complex training dynamics; we therefore use no approximations. As a result, node scores reflect true task performance, and propagating these scores through ancestor statistics keeps the graph's selection policy grounded in reliable---albeit computationally expensive---feedback.

To balance exploration and exploitation, node selection uses a UCT-like score (cf.\ Appendix~\ref{app:mcgs}):
\begin{equation}
    \textsc{UCT}(v_i) = \bar{f}(v_i) \;+\; c(t)\,\sqrt{\frac{\ln N}{n_i}}
    \label{eq:uct-main}
\end{equation}
where $\bar{f}(v_i)$ is the mean reward among $v_i$'s descendants, $N$ is the total number of evaluated nodes, $n_i$ is the visit count for $v_i$, and $c(t)$ is a time-decaying exploration coefficient. Early in the search, a large $c(t)$ encourages broad coverage of the configuration space; as the search progresses $c(t)$ decreases, and the agent transitions to top-$K$ exploitation---expanding only the $K$ highest-scoring nodes, where $K$ is chosen based on the current graph structure and remaining compute budget. The effect is that early iterations explore the configuration space broadly, while later iterations focus compute on refining the most promising configurations.

As independent branches develop, they may discover complementary strategies. The agent can \emph{fuse} the top-$K$ nodes across branches into a single candidate:
\begin{equation}
    \pi_{\mathrm{fused}} = \textsc{Fuse}\!\bigl(\textsc{TopK}(\mathcal{G},\, K)\bigr)
    \label{eq:fuse}
\end{equation}
For example, one branch may discover that adding hard negatives resolves label confusion, while another finds that increasing the number of training epochs improves convergence; fusion produces a configuration that combines both. Fusion also serves as one of two \emph{stagnation recovery} mechanisms. When the agent detects that a branch has stopped improving, it triggers either (i)~\emph{evolution}---a trajectory-aware mutation that tries a fundamentally different approach while preserving what worked---or (ii)~fusion with a complementary branch. Together, these mechanisms prevent the search from becoming trapped in local optima.

\subsection{Data Curation}
\label{sec:data-curation}

Both modes follow the same data curation principles when constructing training datasets. Although the \emph{source} of examples differs (\textit{cold-start mode} relies on downloaded or synthesized data, while \textit{production mode} uses corrected inference failures) the same quality controls are applied in both cases to ensure consistency.

\paragraph{Dataset Composition}

Every training dataset is assembled from three slices, with target proportions that vary by mode and model maturity. \textbf{Gold examples} (40--60\%) are correct input--output pairs sourced from downloaded benchmarks in cold-start mode or corrected failures in production mode. \textbf{Hard negatives} (25--35\%) are confusable inputs where the correct answer differs from a plausible alternative, forcing the model to learn fine-grained decision boundaries. \textbf{Replay buffer} (10--20\%) is sampled from the parent model's training data to prevent catastrophic forgetting \citep{Kirkpatrick2016OvercomingCF}; this slice is used only in production mode when improving an already fine-tuned model and is omitted (with its allocation redistributed to gold examples) when training from a base model.

Empirically, we found that the use of hard negatives is essential. For every gold example at a decision boundary, the agent generates a counterexample with a similar input but a different correct output, teaching the model \emph{when not to fire} rather than only when to fire. The replay buffer prevents catastrophic forgetting when improving an already fine-tuned model; when starting from a base model, the replay slice is reallocated to gold examples.

\paragraph{Quality Controls}

Five constraints govern dataset quality:

\begin{enumerate}
    \item \textbf{The 2-for-1 rule}: For each challenging case, generate one gold example (correct label) and one hard negative (similar input, different correct answer).
    \item \textbf{Label balancing}: No single label exceeds $3\times$ the count of any other label. This prevents the model from learning majority-class priors.
    \item \textbf{Context-length matching}: The length distribution of training examples must match the distribution of realistic inputs (cold-start) or failing inferences (production).
    \item \textbf{Entity diversification} (NER tasks): No single entity value appears more than 2--3 times. Entity values are replaced with synthetic equivalents across examples to prevent memorization of surface forms.
    \item \textbf{Chain-of-thought annotation} (generation tasks): A teacher model (GPT-4.1, DeepSeek-R1) generates step-by-step reasoning chains for training examples, teaching the model \emph{why} an answer is correct rather than only \emph{what} the answer is.
\end{enumerate}

Additionally, each label requires 3--5 distinct surface-text patterns to enforce diversity beyond individual example counts.

\paragraph{Dataset Sizing}

The agent targets compact, high-quality datasets over large noisy ones. For classification and NER tasks, 100--200 total examples is typical; for generation tasks, 500--3,000 examples depending on task complexity. Empirically, the system monitors for accuracy regression when adding data: if expanding the dataset degrades validation-set accuracy, the agent rolls back immediately (see Section~\ref{sec:iteration-policy}).

\subsection{Iteration Policy}
\label{sec:iteration-policy}

Both cold-start and production modes share a common iteration policy that governs how the agent responds to evaluation results after each training run. The agent follows a structured decision tree based on the validation-set score $f(\pi)$:

\begin{enumerate}
    \item \textbf{Score $<$ 0.80}: The dataset needs fundamental rework. The agent analyzes remaining failures, identifies systematic gaps (missing label coverage, distribution mismatch, or insufficient hard negatives), and rebuilds the training data. This threshold treats the problem as a data problem, not a hyperparameter problem.
    \item \textbf{Score 0.80--0.95}: Tune hyperparameters first -- more epochs, different learning rate, larger base model, or adjusted LoRA rank. The dataset is held fixed to isolate the effect of optimization changes.
    \item \textbf{Score $>$ 0.95}: Add 2--3 targeted examples per remaining failure pattern. At this stage, bulk data changes risk regressions, so the agent switches to surgical augmentation.
    \item \textbf{Regression from previous iteration}: Roll back to the previous dataset and configuration immediately. More data is not always better, and the agent treats any score decrease as a signal to revert rather than compensate.
\end{enumerate}

This policy encodes two key principles. First, the dominant source of error shifts as the score improves: a low score signals a data problem, a mid-range score signals an optimization problem, and a high score requires surgical intervention. Second, rollback is always preferred over accumulation -- cascading ``fixes'' that each introduce new failure modes are the primary cause of stagnation in iterative fine-tuning.

\subsection{Cold-Start Mode}

In cold-start mode, the agent receives only a natural-language task description and must produce a fine-tuned model autonomously. Formally, the input is a task specification
\begin{equation}
    \tau = (\text{description},\; \mathcal{M},\; \text{constraints})
    \label{eq:cold-input}
\end{equation}
where $\text{description}$ is a natural-language instruction (e.g., ``fine-tune Llama~3.2-3B on ARC-Challenge''), $\mathcal{M}$ is the base model family, and $\text{constraints}$ captures any user-specified requirements (target metric, output format, domain restrictions). Because no prior model exists, the agent solves the unconstrained form of the optimization objective in Eq.~\ref{eq:objective}:
\begin{equation}
    \pi^* = \arg\max_{\pi \in \Pi}\; f(\pi)
    \label{eq:cold-objective}
\end{equation}
where $f(\pi)$ is the held-out validation-set score of the pipeline $\pi = (D, H, S)$ under the task-appropriate metric. No regression constraint is applied since there is no deployed model to protect. The workflow proceeds in five stages described below.

\paragraph{Input and Tools}

The agent has four tools: \texttt{web\_search} (Exa deep research API), \texttt{bash} (unrestricted shell access with pre-loaded training helpers), \texttt{read\_file}, and \texttt{edit\_file}. The agent can also spawn sub-agents via \texttt{delegate\_task} to parallelize work---for example, delegating data synthesis to one sub-agent while another constructs evaluation sets. Sub-agents inherit the same filesystem, so the main agent can read summaries written to disk by a sub-agent without loading raw data into its own context.

\paragraph{Task Analysis}

Given $\tau$, the agent must determine \emph{what kind of task} this is, \emph{what data exists}, and \emph{what performance is achievable}. The workflow has three stages.

\textbf{(1)} \textit{Task classification:} The agent parses the task description to determine the task type $t$ (\textit{classification}, \textit{NER}, and \textit{generation}) and selects the appropriate model family from $\mathcal{M}$. The task type determines the supervision format (direct labels vs.\ chain-of-thought), the evaluation metric (accuracy, F1, or ROUGE), and the data-curation strategy (Section~\ref{sec:data-curation}).

\textbf{(2)} \textit{Data acquisition:} The agent conducts web research to locate relevant datasets, code examples, and domain knowledge. When the user specifies a known benchmark (e.g., CoNLL-2003, GSM8K), the agent downloads the actual benchmark data rather than generating synthetic substitutes. For custom tasks without public datasets, the agent uses a teacher model to synthesize seed examples, which are then validated and expanded.

\textbf{(3)} \textit{Baseline survey:} The agent surveys published baselines and state-of-the-art (SOTA) results for the target task to calibrate its performance targets. This step identifies known challenges---label ambiguity, class imbalance, domain-specific formatting requirements---that inform the data-curation strategy. When SOTA results are available, the agent sets its initial accuracy target relative to published numbers rather than using a fixed threshold.

\paragraph{Evaluation Setup}

Before any training, the agent constructs a held-out evaluation set $\mathcal{E}$ that defines success criteria. The set is composed of three complementary slices:
\begin{equation}
    \mathcal{E} = \mathcal{E}_{\text{pos}} \cup \mathcal{E}_{\text{neg}} \cup \mathcal{E}_{\text{boundary}}
    \label{eq:cold-eval}
\end{equation}
where $\mathcal{E}_{\text{pos}}$ contains correct input--output pairs covering the full label or entity range, $\mathcal{E}_{\text{neg}}$ contains negative examples (inputs that should \emph{not} trigger any label), and $\mathcal{E}_{\text{boundary}}$ contains confusable pairs at decision boundaries designed to test fine-grained discrimination. For generation tasks, $\mathcal{E}_{\text{neg}}$ is replaced by adversarial inputs (e.g., ill-posed questions, out-of-domain prompts) and $\mathcal{E}_{\text{boundary}}$ contains problems that require multi-step reasoning or edge-case handling. This validation set is held fixed throughout iteration and is never included in the training data.

The default stopping criterion requires:
\begin{equation}
    f(\pi) \geq \tau \quad \text{on } \mathcal{E}, \qquad \tau = 0.96 \text{ by default}
    \label{eq:cold-convergence}
\end{equation}
which mirrors the score gate in production mode (Eq.~\ref{eq:convergence}) but omits the regression constraint since no prior model exists. The default threshold of 0.96 applies to accuracy or F1 for classification/NER and to LLM-as-judge scores for generation tasks. Crucially, this threshold is a default, not a hard constant: the agent can adjust it based on dataset characteristics and base model capabilities. Frontier LLMs are effective at recognizing genuine plateaus---for instance, when remaining failures reflect fundamental model capacity limitations (e.g., a 3B model's inability to memorize factual knowledge) or irreducible label ambiguity rather than fixable training deficiencies. In such cases, the agent lowers the target and terminates early rather than wasting compute on unachievable goals.

\paragraph{Curriculum Synthesis}

The agent assembles training data following the shared quality controls in Section~\ref{sec:data-curation}. Because cold-start mode operates from a base model with $D_{\text{parent}} = \emptyset$, no replay buffer is needed and the resulting dataset is composed of two complementary slices:
\begin{equation}
    D_{\text{cold}} = D_{\text{gold}} \cup D_{\text{hard}}
    \label{eq:curriculum-cold}
\end{equation}
where $D_{\text{gold}}$ consists of correct input--output pairs sourced from downloaded benchmark data or teacher-model synthesis, and $D_{\text{hard}}$ contains hard negatives---confusable inputs where the correct answer differs from a plausible alternative---generated using the 2-for-1 rule (Section~\ref{sec:data-curation}). The replay allocation is redistributed to $D_{\text{gold}}$, yielding target proportions of approximately $|D_{\text{gold}}| : |D_{\text{hard}}| \approx 65{:}35$.

For generation tasks, the agent selects between GPT-4.1 and DeepSeek-R1 as the teacher model. In practice, DeepSeek-R1 is preferred for mathematical and scientific reasoning (e.g., GSM8K, ARC-Challenge), while GPT-4.1 is preferred for code generation and general-knowledge tasks (e.g., HumanEval, TriviaQA).

\paragraph{Training and Iteration}

The agent navigates the pipeline space $\Pi$ using the iterative search procedure described in Section~\ref{sec:search-space}, with full MCGS when the search graph is explicitly maintained. At each iteration, it trains at least two configurations in parallel---for example, full fine-tuning vs.\ LoRA, or different base models---and evaluates each on the full evaluation set $\mathcal{E}$. Let $\pi_i$ denote the pipeline at iteration $i$ with score $f(\pi_i)$. The agent follows the shared iteration policy (Section~\ref{sec:iteration-policy}): dataset rework when $f(\pi_i) < 0.80$, hyperparameter tuning when $0.80 \leq f(\pi_i) < 0.95$, and targeted augmentation when $f(\pi_i) \geq 0.95$. After each iteration, the agent analyzes remaining failures on $\mathcal{E}$ to determine whether the next intervention should target the dataset $D$, the hyperparameters $H$, or the learning strategy $S$ (e.g., switching from direct-answer to chain-of-thought supervision).

If an iteration regresses---$f(\pi_{i+1}) < f(\pi_i)$---the agent rolls back to the previous configuration immediately rather than attempting to compensate. The search terminates when the convergence criterion in Eq.~\ref{eq:cold-convergence} is met or the compute budget of 1{,}500 LangGraph turns is exhausted, whichever comes first.

\subsection{Production Mode}

Production mode is the system's primary contribution. Given a deployed model with real inference failures, the agent autonomously diagnoses, fixes, and verifies improvements. Formally, the input is a deployed model $M_0$ and a set of $n$ judged inference traces
\begin{equation}
    \mathcal{T} = \{(x_i,\; \hat{y}_i,\; y_i^*,\; v_i,\; r_i)\}_{i=1}^{n}
    \label{eq:traces-input}
\end{equation}
where $x_i$ is the input, $\hat{y}_i$ is the model's prediction, $y_i^*$ is the corrected output (from an LLM-as-judge or human reviewer), $v_i \in \{\texttt{pass}, \texttt{fail}\}$ is the verdict, and $r_i$ is the judge's reasoning. The agent's objective is to find a training pipeline $\pi$ that produces an improved model $M_1$ satisfying
\begin{equation}
    f(\pi) > f(\pi_0) \quad \text{on } \mathcal{T}_{\text{fail}}, \qquad
    r(\pi) \leq \epsilon \quad \text{on } \mathcal{T}_{\text{pass}}
    \label{eq:prod-objective}
\end{equation}
where $\pi_0$ denotes the pipeline (if any) that produced $M_0$, $f(\cdot)$ is the validation-set scoring function, and $r(\cdot)$ counts regressions on a held-out set of previously-passing examples.
\begin{figure}[htbp]
\centering
\includegraphics[width=0.95\linewidth]{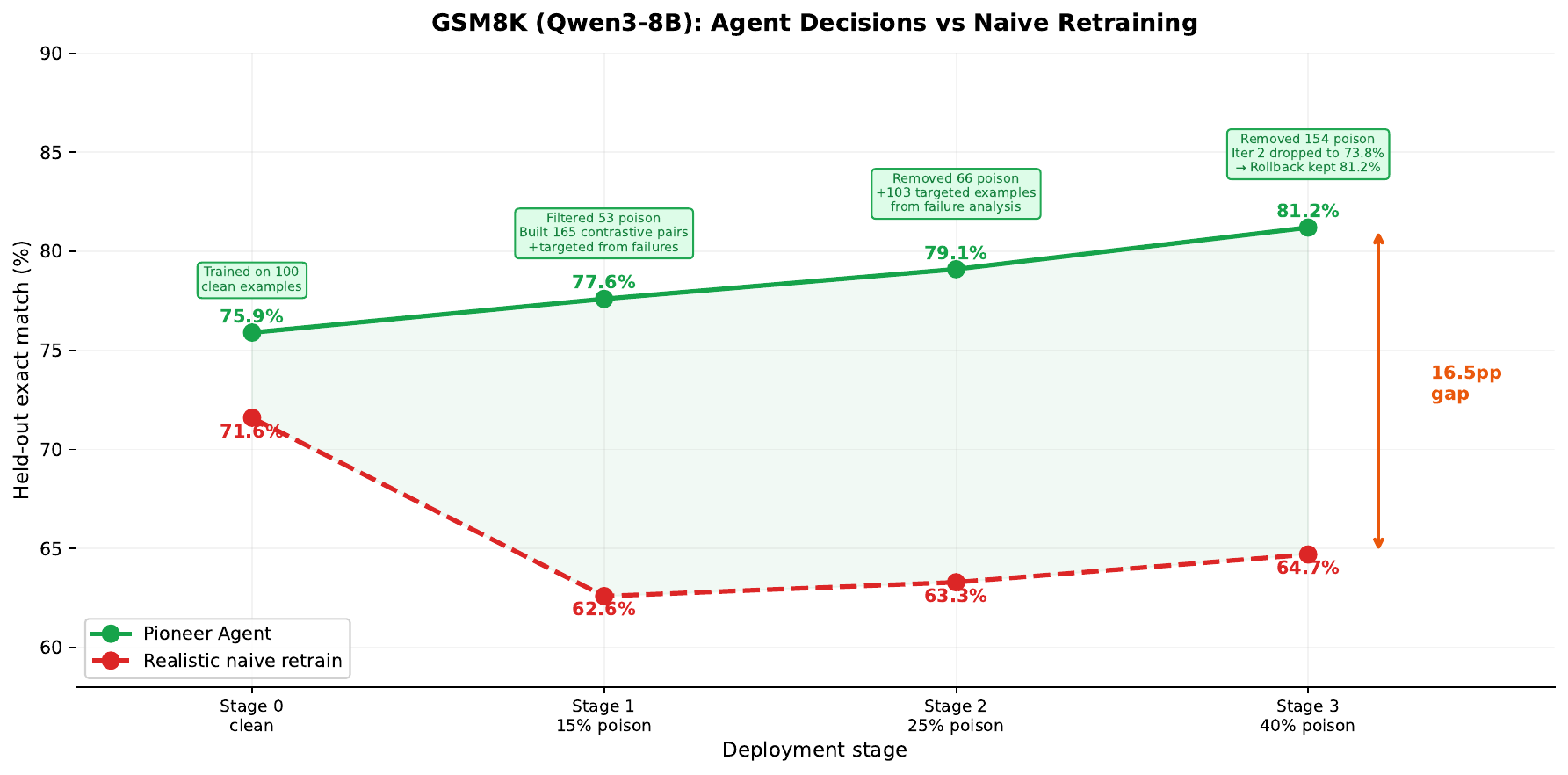}
\caption{\textbf{Annotated stage-based deployment simulation on GSM8K (Qwen3-8B).} At each deployment stage, the annotation shows the agent's key decision: poison filtering, contrastive pair construction, targeted example generation from failure analysis, and automatic rollback when an iteration regresses. Naive retraining trains on all data indiscriminately and degrades sharply on the first noisy stage.}
\label{fig:adaptft-gsm8k-stages}
\end{figure}




\begin{figure}[h]
\centering
\begin{lstlisting}[
  language=Python,
  basicstyle=\ttfamily\scriptsize,
  keywordstyle=\color{blue}\bfseries,
  stringstyle=\color{teal},
  commentstyle=\color{gray}\itshape,
  backgroundcolor=\color{gray!8},
  frame=single,
  rulecolor=\color{gray!40},
  framesep=6pt,
  xleftmargin=8pt,
  xrightmargin=8pt,
  aboveskip=8pt,
  belowskip=8pt,
  breaklines=true,
  columns=fullflexible,
  showstringspaces=false,
  caption={Example \texttt{query\_traces} call combining SQL with bash post-processing in a single tool invocation. The SQL result is piped to the bash command; only the printed summary enters the agent's context while the full dataset is saved to disk.},
  label=lst:query-traces
]
query_traces(
  sql_query = """
    SELECT input, metadata
    FROM inferences
    WHERE llmaj_verdict = 'fail'
    LIMIT 10000
  """,
  bash_pipeline = """
    python3 -c '
    import sys, json
    rows = json.loads(sys.stdin.read())

    # Group failures by perturbation type
    by_type = {}
    for r in rows:
        for p in r["metadata"].get("perturbations_applied", []):
            by_type.setdefault(p, []).append(r)

    # Print compact summary -> agent context
    for k, v in sorted(by_type.items(), key=lambda x: -len(x[1])):
        print(f"{k}: {len(v)} failures")

    # Save full dataset to disk -> training pipeline
    with open("/tmp/failures.jsonl", "w") as f:
        for r in rows:
            f.write(json.dumps(r) + chr(10))
    print(f"Saved {len(rows)} rows to /tmp/failures.jsonl")
    '
  """
)
\end{lstlisting}
\vspace{-2.5em}
\end{figure}


\paragraph{Input and Tools} The agent has access to five tools: \texttt{query\_traces} (SQL with optional bash post-processing), \texttt{trace\_analysis\_subagent} (delegated complex log analysis), \texttt{bash} (unrestricted shell with pre-loaded training helpers), \texttt{read\_file}, and \texttt{edit\_file}. More specifically, the \texttt{query\_traces} tool accepts both a SQL query and an optional \texttt{bash\_pipeline} command, piping the query result directly to the bash command's stdin in a single invocation. This design allows the agent to filter, aggregate, and persist large result sets on disk without loading raw rows into its context window. In practice, the agent's context rarely contains more than ${\sim}50$ representative examples even when operating over databases with tens of thousands of rows.


\paragraph{Failure Diagnosis}

Given a set of judged inference traces, the agent must decide \emph{what is broken}, \emph{why it is broken}, and \emph{what can be fixed by training}. We break this into four stages: (1)~\emph{trace ingestion}, which partitions the logs into failure and passing sets; (2)~\emph{taxonomy construction}, which clusters failures into actionable categories and labels each as fixable or external; (3)~\emph{live confirmation}, which probes the deployed model to verify that identified weaknesses are systematic; and (4)~\emph{parent model awareness}, which inspects model lineage to determine whether corrective data should complement an existing training set or be built from scratch. We describe each stage in turn below.

\textbf{(1)} \textit{Trace ingestion:} The agent operates over the production inference database described above. Each record is a tuple
\begin{equation}
    t_i = (x_i,\; \hat{y}_i,\; y_i^*,\; v_i,\; r_i,\; m_i)
    \label{eq:trace}
\end{equation}
where the first five fields are as defined in Eq.~\ref{eq:traces-input} and $m_i$ encodes judge metadata (the judge model, prompt template, and evaluation criteria). The judge itself may be a deterministic scorer---such as a token-level F1 function that computes overlap against a gold reference---or an LLM judge such as DeepSeek; in the deterministic case, $r_i$ contains the score breakdown and $y_i^*$ reduces to the gold reference. When a human override is present, it supersedes the automated verdict. The agent queries these records via SQL with automatic row-level filtering to the current user's data, obtaining the failure and passing sets
\begin{equation}
    \mathcal{T}_{\text{fail}} = \{t_i \mid v_i = \texttt{fail}\}, \qquad
    \mathcal{T}_{\text{pass}} = \{t_i \mid v_i = \texttt{pass}\}.
    \label{eq:trace-sets}
\end{equation}
This signal tells the agent which inferences failed, why they failed, and what the corrected output should be.

\textbf{(2)} \textit{Taxonomy construction:} The agent partitions $\mathcal{T}_{\text{fail}}$ into $K$ failure clusters
\begin{equation}
    \{C_1, \dots, C_K\} \quad \text{such that} \quad \bigcup_{k=1}^{K} C_k = \mathcal{T}_{\text{fail}}.
    \label{eq:clusters}
\end{equation}
For each cluster $C_k$, the agent records its size $|C_k|$, the dominant input characteristics (length distribution, structural patterns, entity-type composition), and a fixability label $\phi_k \in \{\texttt{fixable}, \texttt{external}\}$ indicating whether the failures are addressable through training or are instead attributable to external factors such as prompt-design errors, schema mismatches, or genuinely ambiguous inputs. Complex analysis is delegated to a \emph{Trace Analyzer} sub-agent (with $10\times$ higher token limits), which produces structured outputs like failure taxonomies, confusion matrices, and representative input examples, which are saved to disk. The main agent reads only the summary, preserving context for high-level decision-making.

\textbf{(3)} \textit{Live confirmation:} Before constructing training data, the agent first verifies that each identified weakness is systematic rather than an artifact of sampling noise. To do this, for each fixable cluster $C_k$ with $\phi_k = \texttt{fixable}$, the agent synthesizes a probe set $P_k = \{p_1, \dots, p_n\}$ of targeted inputs designed to trigger the hypothesized failure mode, and evaluates the deployed model on $P_k$.

Crucially, probe construction depends on the specific failure mode: if the model confuses two labels, the agent generates boundary-case inputs that distinguish them; if long inputs cause failures, it extends previously correct short inputs to identify the breaking point; and if specific entity types are missed, it creates synthetic examples that explicitly include those entities.

In practice, this step serves two roles: confirming that the weakness is systematic and generating targeted failing examples that are directly used to build the corrective training dataset.

\textbf{(4)} \textit{Parent model awareness:} 
To ensure targeted improvement, the agent first determines what supervision the model has already received. It inspects the \texttt{model\_id} associated with failing inferences to recover the model's training lineage, distinguishing between base checkpoints and previously fine-tuned models.

Let $D_{\text{parent}}$ denote the training dataset from the model's most recent fine-tuning run, with $D_{\text{parent}} = \emptyset$ when the deployed model is a base checkpoint. If the model is a base model (e.g., \texttt{fastino/gliner2-base-v1}), no prior task-specific supervision exists, so the agent builds the training dataset from scratch, using inference logs only to identify failure patterns. If the model is already fine-tuned (identified by a UUID), the agent retrieves $D_{\text{parent}}$ and constructs a \emph{complementary} dataset that targets uncovered failure modes rather than duplicating existing supervision. To reduce catastrophic forgetting, the agent also includes a replay buffer
\begin{equation}
D_{\text{replay}} \subset D_{\text{parent}}, \qquad
|D_{\text{replay}}| \approx (0.1\text{--}0.2)\,|D_{\text{parent}}|,
\label{eq:replay}
\end{equation}
which helps preserve previously learned behaviors while adapting the model to new errors. In all cases, training restarts from the base foundation model rather than a previously fine-tuned checkpoint. This ensures that the model’s behavior is fully determined by the current iteration’s dataset, rather than a mixture of accumulated updates from earlier runs. As a result, failures can be directly attributed to gaps or errors in the current dataset, and rollback becomes straightforward: reverting to a previous dataset cleanly restores the corresponding model behavior without needing to untangle interactions between successive fine-tuning steps.

\begin{figure}
\centering
\includegraphics[width=0.98\linewidth]{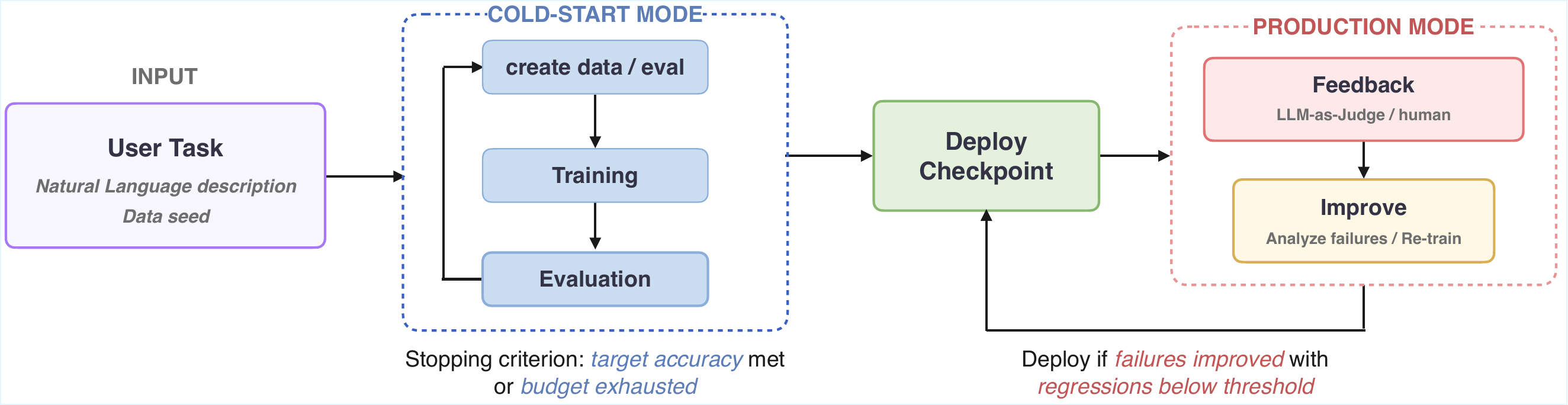}
\caption{\textbf{End-to-end closed-loop pipeline unifying both Pioneer Agent modes.} In cold-start mode (left), the agent progresses from a natural-language task description through data acquisition, evaluation setup, curriculum synthesis, and agent-guided iteration. In production mode (right), judged inference failures feed into failure diagnosis, taxonomy construction, targeted curriculum synthesis, retraining with regression gating, and conditional deployment. The two modes share the same data-curation principles (Section~\ref{sec:data-curation}) and iteration policy (Section~\ref{sec:iteration-policy}), enabling a single system to handle both initial model creation and continuous post-deployment improvement.}
\label{fig:closed-loop}
\end{figure}

\paragraph{Evaluation Setup}

The evaluation relies on two complementary sets derived from $\mathcal{T}$. The \emph{evaluation set} comprises the entirety of $\mathcal{T}_{\text{fail}}$ evaluated using the same judgment method that produced the original verdicts (a deterministic scorer or an LLM judge with identical criteria stored in inference table metadata). The \emph{regression set} is a structured sample of previously passing inferences, constructed to cover the model’s existing strengths across relevant dimensions (e.g., label and entity types for classification/NER).

Let $a(\pi)$ denote the evaluation-set score and $r(\pi)$ the regression count of pipeline~$\pi$. The default stopping criterion requires both conditions to hold simultaneously:
\begin{equation}
a(\pi) \geq \tau, \qquad r(\pi) \leq \epsilon, \qquad \tau = 0.96 \text{ by default}
\label{eq:convergence}
\end{equation}
where $\epsilon = 2$ as defined in Eq.~\ref{eq:objective}. As in cold-start mode, 0.96 is the default threshold but the agent may lower it when it determines that remaining failures are not addressable by training---for example, when the failure taxonomy reveals that the dominant error cluster consists of adversarial inputs, out-of-distribution queries, or upstream label errors. The agent diagnoses these cases by analyzing failure patterns programmatically and reasoning about whether additional training data could plausibly resolve them. A candidate model is accepted only when it satisfies both gates; otherwise, the agent continues iterating or rolls back to the previous best checkpoint.

\paragraph{Curriculum Synthesis}

From the confirmed fixable clusters ${C_k \mid \phi_k = \texttt{fixable}}$, the agent constructs a targeted training curriculum following the shared curation principles in Section~\ref{sec:data-curation}. The resulting dataset is composed of three complementary slices:
\begin{equation}
D_{\text{post}} = D_{\text{gold}} \cup D_{\text{hard}} \cup D_{\text{replay}}
\label{eq:curriculum-prod}
\end{equation}
where $D_{\text{gold}}$ consists of corrected versions of failed traces (using judge or human annotations $y_i^*$), $D_{\text{hard}}$ contains hard negatives derived from confusion pairs—inputs for which $M_0$ produces plausible but incorrect outputs—and $D_{\text{replay}}$ is the replay buffer defined in Eq.~\ref{eq:replay}, included to preserve previously learned behavior. The relative composition ($|D_{\text{gold}}| : |D_{\text{hard}}| : |D_{\text{replay}}|$) is adapted to the observed failure modes. More specifically, clusters dominated by recall errors are addressed by increasing the proportion of gold examples, while clusters characterized by precision errors (e.g., false positives or label confusion) receive a higher proportion of hard negatives.

Furthermore, all examples are constrained by the quality controls described in Section~\ref{sec:data-curation}, ensuring that the resulting dataset is both targeted to the observed failures and consistent with the overall data curation framework.

\paragraph{Training and Iteration}

The agent trains on $D_{\text{post}}$ and evaluates the resulting model on both the evaluation set and the regression set. If $r(\pi) \geq \epsilon$, the iteration is rejected and the agent reverts to the previous best checkpoint. Iterations follow the shared policy in Section~\ref{sec:iteration-policy}: dataset rework below 80\% score, hyperparameter tuning between 80--95\%, and targeted augmentation above 95\%. Meanwhile, the agent can run up to 500 LangGraph turns per production run, using sub-agents for data-heavy steps such as failure analysis and trace sampling. When the stopping criterion in Eq.~\ref{eq:convergence} is met, the improved model $M_1$ is promoted as the new deployment candidate.

\subsection{Structural Safeguards}

A central risk in autonomous fine-tuning is overfitting to production inference data, which is often noisy, biased, and temporally correlated. The mechanisms described above—hard negatives and label balancing (Section~\ref{sec:data-curation}), rollback-first iteration (Section~\ref{sec:iteration-policy}), parallel training of 2--3 configurations, and regression gating in production mode—jointly mitigate this risk by constraining both data composition and update acceptance. One additional mechanism completes the picture.

\textbf{Cross-checkpoint regression gate.} In production mode, the evaluation set evolves over time as new failures are discovered, which creates a risk of \emph{temporal overfitting} \citep{QuioneroCandela2009DatasetSI, Widmer1996LearningIT, gama2014conceptdrift, bayram2022modeldegradation}: a model may improve on the latest failure slice while silently degrading on previously fixed behaviors. To prevent this, once a candidate model reaches the score threshold on the current evaluation set, it is additionally evaluated on the evaluation set from the immediately preceding checkpoint. The model must satisfy $r(\pi) \leq \epsilon$ (at most 2 regressions; Eq.~\ref{eq:objective}) on \emph{both} sets. This enforces cross-iteration consistency by requiring that improvements on newly introduced data do not come at the expense of prior gains. Combined with the replay buffer, which re-exposes earlier training signal, this produces a ratchet effect: each accepted update must preserve previously validated behavior while incorporating new improvements, yielding monotonically non-decreasing performance across successive fine-tuning rounds.

\begin{table}
\centering
\small
\setlength{\tabcolsep}{4pt}
\renewcommand{\arraystretch}{1.1}

\begin{tabular}{@{}llllrrr@{}}
\toprule
\rowcolor{headergray}
Scenario & Task & Base Model & Domain & Fail & Pass & Tier \\
\midrule

\rowcolor{sectiongray}
\multicolumn{7}{@{}l}{\textit{Benchmark-derived synthetic-log scenarios}} \\

GSM8K        & Math generation   & Qwen3-8B     & Math reasoning     & \cellcolor{lightred}751 & \cellcolor{lightgreen}577 & Hard   \\
GSM8K        & Math generation   & Llama 3.2-3B & Math reasoning     & \cellcolor{lightred}903 & \cellcolor{lightgreen}425 & Hard   \\
ARC-Challenge & Multiple choice  & Llama 3.2-3B & Science reasoning  & \cellcolor{lightred}765 & \cellcolor{lightgreen}735 & Hard   \\
TriviaQA     & Open QA           & Llama 3.2-3B & Trivia knowledge   & \cellcolor{lightred}798 & \cellcolor{lightgreen}702 & Hard   \\
HumanEval    & Code completion   & Qwen3-8B     & Python functions   & \cellcolor{lightred}395 & \cellcolor{lightgreen}235 & Medium \\
XSum         & Summarization     & Qwen3-8B     & News articles      & \cellcolor{lightred}${\sim}612$             & \cellcolor{lightgreen} ${\sim}438$             & Medium \\
SAMSum       & Summarization     & Qwen3-8B     & Dialogue           & \cellcolor{lightred}${\sim}587$             & \cellcolor{lightgreen} ${\sim}463$             & Medium \\

\midrule

\rowcolor{sectiongray}
\multicolumn{7}{@{}l}{\textit{Production-style deployment case studies}} \\

CLINC150     & 30-class classif. & GLiNER2-base & Intent detection   & \cellcolor{lightred}453  & \cellcolor{lightgreen}2,546 & Medium \\
CoNLL-2003   & NER               & GLiNER2-base & Entity extraction  & \cellcolor{lightred}2,740 & \cellcolor{lightgreen}280   & Hard   \\

\bottomrule
\end{tabular}

\caption{\textbf{AdaptFT-Bench scenarios used in this paper.} The first seven are benchmark-derived synthetic-log scenarios evaluated with the stage-based protocol below; CLINC150 and CoNLL-2003 are production-style deployment case studies.}
\label{tab:adaptft-scenarios}
\vspace{-1em}
\end{table}

 \textbf{Confidence calibration.} In production settings, model confidence is often misaligned with true correctness, especially under distribution shift \citep{Guo2017OnCO}: models tend to be overconfident on systematic failure modes. Left uncorrected, this reduces the effectiveness of human-in-the-loop annotation, as high-confidence errors are less likely to be reviewed. To address this, the adaptive annotation system tracks historical accuracy per label and adjusts raw model confidence: $\textit{calibrated} = \textit{weight} \times \textit{actual\_accuracy} + (1-\textit{weight}) \times \textit{raw\_confidence}$. A model that is overconfident on a label it historically gets wrong will have its confidence pulled down. TF-IDF similarity \citep{Salton1988TermWeightingAI} identifies related unlabeled items when a human corrects a prediction, propagating corrections to similar examples and preventing the same incorrect pattern from recurring.

These safeguards were validated empirically. On the CLINC150 production improvement run (Section~\ref{sec:clinc150}), the agent's V2 iteration regressed from 99.3\% to 98.5\% -- and the system correctly rolled back to V1 rather than continuing to iterate. On SMS Spam (Section~\ref{sec:sms-spam}), the recall-precision balancing pattern demonstrated the hard negative safeguard in action: each iteration of adding spam examples to fix recall introduced new false positives, which the regression gate caught, triggering a corrective iteration of hard ham examples.

\section{AdaptFT-Bench}
\label{sec:adaptft}

Existing ML benchmarks evaluate model performance on static test sets. No standard benchmark exists for evaluating \emph{agents that improve models from production failures}. AdaptFT-Bench fills this gap by providing reproducible scenarios where an agent must analyze a deployed model's failures, build training data, retrain, and verify improvements -- all autonomously. The benchmark answers a single question: \textbf{given a deployed model with a known set of failures, can an agent autonomously fix them without breaking what already works?}

AdaptFT-Bench has two components. The primary benchmark consists of seven benchmark-derived synthetic inference scenarios built from public datasets using our perturbation pipeline (Appendix~\ref{app:noise}). In addition, we report two staged deployment case studies (CLINC150 and CoNLL-2003), in which public benchmark data was deployed through the Pioneer Agent platform and the resulting inference logs with LLM-as-judge verdicts were used as input to the production improvement loop. Full benchmark construction methodology --- including scenario selection, synthetic inference log construction, failure and regression set construction, difficulty tiers, and anonymization procedures --- is provided in Appendix~\ref{app:adaptft-details}.

\subsection{Stage-Based Evaluation Protocol}

AdaptFT-Bench has two evaluation regimes. The seven benchmark-derived scenarios (two GSM8K runs \cite{cobbe2021training}, ARC-Challenge \cite{clark2018think}, TriviaQA \cite{joshi2017triviaqa}, HumanEval \cite{chen2021evaluating}, XSum \cite{narayan2018dont}, and SAMSum \cite{gliwa2019samsum}) use synthetic inference logs generated from benchmark data with our noise injection pipeline (Appendix~\ref{app:noise}). CLINC150 \cite{larson2019evaluation} and CoNLL-2003 \cite{tjong2003introduction} use production-style deployment logs and are reported separately as case studies rather than inference simulations. For the stage-based scenarios, logs are organized into three deployment stages with increasing poison rates (15\% $\rightarrow$ 25\% $\rightarrow$ 40\%). Each stage contributes roughly 500 new inference logs and is split 70/30 into train/test. The held-out evaluation set is the union of all stage test splits, so every checkpoint is measured on the same mixed clean+noisy slice.

We distinguish \emph{fixable noise} from \emph{poisonous noise}. Fixable noise includes typos, misspellings, grammatical corruption, casing errors, truncation, preamble injection, and code-switching -- perturbations where the original intent and answer remain valid. Poisonous noise includes false premises, label flips, off-domain queries, prompt injection, jailbreaks, gibberish, and empty inputs -- cases where training on the raw example would actively teach the wrong behavior.

\begin{table}
\centering
\small
\setlength{\tabcolsep}{5pt}
\renewcommand{\arraystretch}{1.0}

\begin{tabular*}{\linewidth}{@{\extracolsep{\fill}} l l r r r r @{}}
\toprule
\rowcolor{headergray}
\textbf{Benchmark} & \textbf{Model} & \textbf{Baseline} & \textbf{Fine-tuned} & \textbf{$\Delta$} & \textbf{Iters} \\
\midrule

\multicolumn{6}{@{}l}{\textit{Decoder models (generation)}} \\

ARC-Challenge  & Llama 3.2-3B & 5.3\% & \textbf{72.6\%} & \textbf{+67.3} & 11 \\
GSM8K          & Llama 3.2-3B & $\sim$8\%\textsuperscript{*} & \textbf{43.7\%} & \textbf{+35.7} & 10 \\
TriviaQA       & Llama 3.2-3B & $\sim$0\%\textsuperscript{*} & \textbf{48.6\%} & \textbf{+48.6} & 9  \\

\addlinespace[3pt]

ARC-Challenge  & Qwen3-8B     & 91.7\% & \textbf{93.3\%} & +1.6  & 13 \\
HumanEval      & Qwen3-8B     & 71.3\% & \textbf{92.7\%} & \textbf{+21.4} & 4  \\
XSum           & Qwen3-8B     & R2: 6.9 & \textbf{17.9} & \textbf{+159\%} & 13 \\
SAMSum         & Qwen3-8B     & R2: 13.4 & \textbf{25.4} & \textbf{+90\%}  & 11 \\

\midrule

\multicolumn{6}{@{}l}{\textit{Encoder model (classification)}} \\

SMS Spam (UCI) & GLiNER2-base & F1: 0.159 & \textbf{0.997} & \textbf{+83.8} & 10 \\

\bottomrule
\end{tabular*}

\caption{
\textbf{Cold-start benchmark results across eight tasks.}
The agent receives only a natural-language task name and autonomously downloads data, constructs train/test splits, synthesizes training curricula, and iterates.
Baselines are zero-shot on the same held-out validation set.
Improvements range from +1.6 pp (Qwen3-8B on ARC, where the base model already scores 91.7\%) to +83.8 pp F1 (SMS Spam).
\textsuperscript{*} Llama~3.2-3B is a base model that cannot follow standard output formats zero-shot; GSM8K baseline is $\sim$8\% with 5-shot prompting.
}
\label{tab:coldstart-results}
\end{table}

In the synthetic benchmark, these labels are known from the perturbation pipeline. In production mode, the agent itself classifies each failing inference as fixable or poisonous during the failure-taxonomy Step (Section~\ref{sec:clinc150}), using the LLM's judgment about whether the original intent and correct answer are recoverable from the corrupted input. Pioneer Agent filters poisonous examples, builds contrastive pairs from fixable noise, trains on the accumulated curated data, and is allowed up to two train $\rightarrow$ evaluate iterations per stage with rollback. This means the deployed adaptive curve is monotone or flat by design: if a stage-specific retrain regresses, the system keeps the previous checkpoint. By contrast, naive retraining always deploys the single checkpoint produced from the latest noisy stage.

Because AdaptFT-Bench evaluates on held-out synthetic inference logs rather than the canonical benchmark test sets, the absolute values in the results tables should not be compared directly to the cold-start results. AdaptFT-Bench measures trend under accumulating production noise and the final gap to naive retraining, not clean benchmark accuracy.

\begin{table}
\centering
\small
\setlength{\tabcolsep}{6pt}
\renewcommand{\arraystretch}{1.2}
\rowcolors{2}{white}{gray!20}
\begin{tabularx}{\linewidth}{@{}>{\raggedright\arraybackslash}p{3.2cm}X@{}}
\toprule
\textbf{Benchmark} & \textbf{Key Finding} \\
\midrule
ARC-Challenge (Llama 3B) & Chain-of-thought training was the largest observed lever (+21pp); DeepSeek-R1 traces outperformed GPT-4.1. Post-peak iterations with more data, higher LoRA rank, or lower learning rate all regressed. \\
GSM8K (Llama 3B) & Fewer epochs (2) with larger batch size (16) were optimal; the model overfits quickly on math patterns. Final accuracy (43.7\%) is competitive for 3B-scale models. \\
HumanEval (Qwen 8B) & Cross-benchmark generalization: trained on MBPP (374 problems), evaluated on HumanEval (164). Adding GPT-4.1-generated solutions \emph{reduced} performance (96.9\%$\to$94.5\%), indicating that external outputs can dilute the training signal. \\
XSum (Qwen 8B) & Summarization overfits rapidly---1 epoch is optimal. Fine-tuning corrected verbosity (32.6-word $\to$ 21-word summaries). \\
SAMSum (Qwen 8B) & System prompt design was the primary lever: a constrained prompt (``1--3 sentences, main outcome, third person, brief'') outperformed all hyperparameter changes. \\
SMS Spam (GLiNER2) & 55 targeted examples added to a 4,458-example base (+1.2\%) increased F1 from 0.9834 to 0.9967 via iterative precision--recall balancing. \\
\bottomrule
\end{tabularx}
\caption{\textbf{Key findings per cold-start benchmark.} Each row highlights the most significant insight from the agent's iteration trajectory. Full per-iteration tables appear in Appendix~\ref{app:cold-start-trajectories}.}
\label{tab:cold-start-insights}
\end{table}

\section{Experiments}

\begin{figure}
\centering
\includegraphics[width=0.98\linewidth]{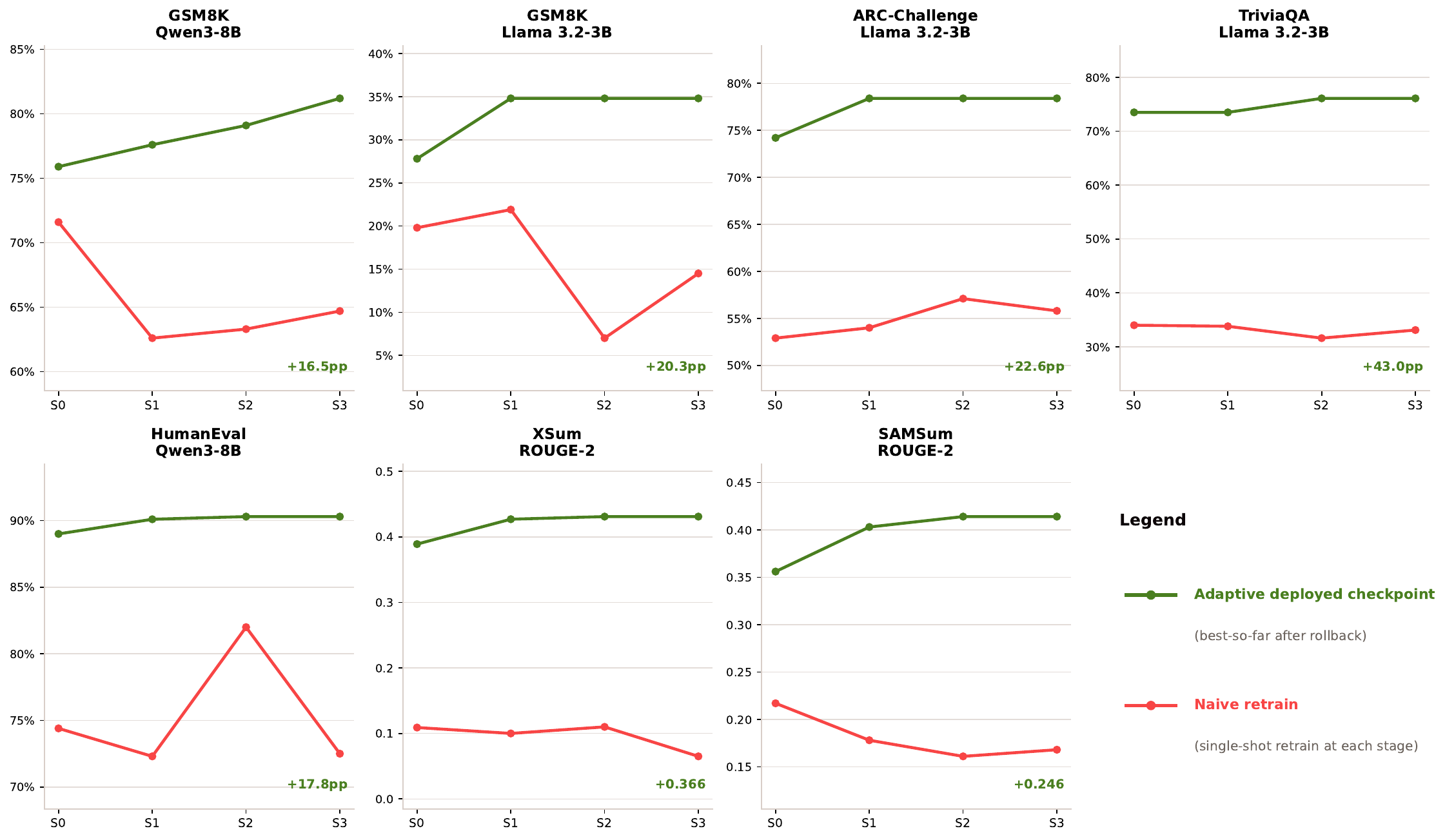}
\caption{\textbf{Stage trajectories across the seven AdaptFT-Bench scenarios.} The deployed adaptive model (green) moves up or stays flat; naive retraining (red) trends down as poisoned logs accumulate.}
\label{fig:adaptft-stage-trends}
\end{figure}

\subsection{Experimental Setup}
\label{sec:setup}

\paragraph{Benchmark Suite} We evaluate Pioneer Agent across nine benchmark scenarios under two settings --- cold-start and production improvement --- designed to capture both initial task bootstrapping and post-deployment adaptation. The evaluation spans a diverse set of task families, including reasoning (ARC-Challenge), mathematical problem solving (GSM8K), open-domain knowledge QA (TriviaQA), code generation (HumanEval), abstractive summarization (XSum, SAMSum), classification (SMS Spam, CLINC150), and named entity recognition (CoNLL-2003). These scenarios cover both encoder and decoder model families (GLiNER2, Llama, and Qwen), and include a mix of base-model initialization and already fine-tuned production systems. Metrics are task-specific and reflect standard evaluation protocols for each domain (e.g., exact match, pass@1, ROUGE, and F1). Table~\ref{tab:adaptft-scenarios} summarizes the full benchmark suite including models and datasets used throughout our experiments.


\paragraph{Evaluation Protocols} We evaluate Pioneer Agent under two protocols corresponding to its two operational modes:

\textbf{(1)} \textit{Cold-start protocol:} The agent is given only a natural-language task specification (e.g., ``fine-tune Llama~3.2-3B on ARC-Challenge'') with no pre-curated dataset or training recipe. It must autonomously acquire data, construct a held-out evaluation set \emph{before} training, synthesize a training dataset, select configurations, and iteratively train and evaluate against the fixed validation set (Section~\ref{sec:pioneer-agent}, Cold-Start Mode). All results are reported on held-out validation sets defined prior to training.

\textbf{(2)} \textit{Production protocol:} The agent is given a deployed model with judged inference traces rather than a clean dataset. It must identify failure patterns, construct a failure taxonomy, synthesize a targeted training curriculum (corrected examples, hard negatives, replay), retrain models, and validate improvements against both a failure set and a regression set of previously correct behavior (Section~\ref{sec:pioneer-agent}, Production Mode).  Production evaluation uses AdaptFT-Bench (Section~\ref{sec:adaptft}), where synthetic inference logs are generated via a noise injection pipeline (Appendix~\ref{app:noise}) and organized into three stages with increasing poison rates (15\% $\rightarrow$ 25\% $\rightarrow$ 40\%). Two additional scenarios (CLINC150 and CoNLL-2003) use production-style deployment logs; full details are in Appendix~\ref{app:adaptft-details}.


\paragraph{Baseline Methods}

We report two baseline families for the production evaluation, capturing both controlled upper bounds and realistic deployment conditions.

\textbf{(1)} \textit{Controlled upper-bound baselines (Table~\ref{tab:adaptft-heldout}).}  
For the held-out poison experiment on GSM8K/Qwen, we retain the original comparison set: naive retraining with oracle corrections, random augmentation, and clean-only training. These baselines isolate the value of filtering and curriculum construction under ideal supervision, where all training labels are correct. The \textbf{naive retrain} baseline trains once on all incoming inference data, including poisoned examples, using oracle-corrected outputs. \textbf{Random augmentation} follows the same procedure but augments failure cases with randomly sampled clean benchmark examples rather than a targeted curriculum. Finally, \textbf{clean-only} training uses exclusively clean benchmark data, with no exposure to noisy deployment logs.

\textbf{(2)} \textit{Realistic naive retrain (Tables~\ref{tab:adaptft-stage-summary}--\ref{tab:adaptft-gsm8k-stages}, Figure~\ref{fig:adaptft-stage-trends}).}  
To reflect real deployment conditions, we introduce a more realistic naive baseline that mirrors common production pipelines without intelligent data curation. All incoming inference data is used for training without any poison filtering. Approximately 30\% of noisy failures receive no human correction, causing the model to retrain on its own incorrect outputs. Among poisonous examples, around 50\% receive incorrect corrections, where annotators answer the corrupted input, while roughly 20\% receive correct corrections by chance. Clean examples always receive correct labels. Training is performed once per deployment stage, with no iteration or rollback.


\subsection{Cold-Start Results}

We evaluate cold-start mode on eight benchmarks spanning reasoning, math, code generation, summarization, and classification (Figure~\ref{fig:benchmark-results}). For each, the agent receives only a task name and must autonomously acquire data, construct evaluation sets, synthesize training curricula, and iterate. Improvement magnitude tracks how far the base model is from task requirements, ranging from +1.6pp on near-saturated baselines to +83.8pp F1 where no prior supervision exists.

The largest gains occur when the base model lacks the required output format or reasoning strategy. On ARC-Challenge (Llama~3.2-3B), the base model scores 5.3\% because it fails to follow the multiple-choice format, generating free-form continuations or new questions instead of selecting an answer; the agent improves performance to 72.6\% (+67.3pp) by discovering chain-of-thought supervision with DeepSeek-R1 traces and fusing complementary search branches. TriviaQA improves from effectively 0\% to 48.6\%, after which the agent attributes remaining errors to the knowledge limits of a 3B model. On GSM8K, it identifies a distinct optimization regime where fewer epochs and larger batch sizes yield the best result (43.7\%, +35.7pp).

When the base model is already capable, the agent shifts from teaching core skills to refining behavior. HumanEval reaches 92.7\% pass@1 (+21.4pp) in 4 iterations via transfer from MBPP. Summarization shows task-specific dynamics: XSum overfits after one epoch, while SAMSum is driven primarily by prompt design. SMS Spam illustrates the high-accuracy regime—after reaching F1~0.98, three rounds adding only 55 targeted examples push F1 to 0.997 via precision--recall balancing. When the baseline is strong (Qwen3-8B on ARC, 91.7\%), gains are modest (+1.6pp) and require recovering from an initial performance dip.

Across benchmarks, the first iteration captures only 40--70\% of the final improvement; remaining gains come from diagnosing failure modes and applying targeted interventions. Table~\ref{tab:cold-start-insights} summarizes the dominant finding per benchmark; full trajectories appear in Appendix~\ref{app:cold-start-trajectories}.

\begin{table}[t]
\centering
\renewcommand{\arraystretch}{1.2}
\setlength{\tabcolsep}{6pt}

\begin{tabular}{@{}ll>{\bfseries}lrrrrr@{}}
\toprule
\rowcolor{headergray}
Benchmark & Model & Metric & Agent C1 & Agent Best & Naive C1 & Naive C4 & Final Gap \\
\midrule

GSM8K & Qwen3-8B & Acc. 
& \cellcolor{lightblue}75.9\% 
& \cellcolor{lightgreen}\textbf{81.2\%} 
& 71.6\% & 64.7\% 
& \textbf{+16.5\%} \\

GSM8K & Llama 3.2-3B & Acc. 
& \cellcolor{lightblue}27.8\% 
& \cellcolor{lightgreen}\textbf{34.8\%} 
& 19.8\% & 14.5\% 
& \textbf{+20.3\%} \\

ARC-Challenge & Llama 3.2-3B & Acc. 
& \cellcolor{lightblue}74.2\% 
& \cellcolor{lightgreen}\textbf{78.4\%} 
& 52.9\% & 55.8\% 
& \textbf{+22.6\%} \\

TriviaQA & Llama 3.2-3B & Acc. 
& \cellcolor{lightblue}73.5\% 
& \cellcolor{lightgreen}\textbf{76.1\%} 
& 34.0\% & 33.1\% 
& \textbf{+43.0\%} \\

HumanEval & Qwen3-8B & pass@1 
& \cellcolor{lightblue}89.0\% 
& \cellcolor{lightgreen}\textbf{90.3\%} 
& 74.4\% & 72.5\% 
& \textbf{+17.8\%} \\

XSum & Qwen3-8B & ROUGE-2 
& \cellcolor{lightblue}38.9 
& \cellcolor{lightgreen}\textbf{43.1} 
& 10.9 & 6.5 
& \textbf{+36.6} \\

SAMSum & Qwen3-8B & ROUGE-2 
& \cellcolor{lightblue}35.6 
& \cellcolor{lightgreen}\textbf{41.4} 
& 21.7 & 16.8 
& \textbf{+24.6} \\

\bottomrule
\end{tabular}

\caption{\textbf{Production results across AdaptFT-Bench scenarios.} 
Agent columns show the best deployed checkpoint with rollback, while naive columns correspond to single-shot retraining per stage. 
Colored cells highlight Agent performance (blue = initial, green = best).}
\label{tab:adaptft-stage-summary}
\end{table}

\subsection{Production Mode Results}

We now evaluate Pioneer Agent's production mode on the core challenge of this work: improving deployed models from noisy, real-world feedback rather than clean, static datasets. Production systems must adapt under distribution shift, imperfect or inconsistent annotations, and adversarial inputs, while preserving previously correct behavior. To stress-test these capabilities under controlled conditions, we use AdaptFT-Bench, a stage-based benchmark with synthetic inference logs that simulate accumulating noise across seven scenarios (Section~\ref{sec:adaptft}).

Table~\ref{tab:adaptft-stage-summary} summarizes results across the seven stage-based AdaptFT-Bench scenarios. Each scenario simulates a deployed model receiving three waves of production traffic with increasing poison rates (15\% $\rightarrow$ 25\% $\rightarrow$ 40\%), and the table reports both the agent's first checkpoint (C1, trained on clean data only) and its best deployed checkpoint after all three stages, alongside the naive baseline at the same points. The ``Final Gap'' column captures the cumulative advantage of adaptive curation over indiscriminate retraining by the end of the simulation.

\subsubsection{Main results.} Across all seven scenarios, Pioneer Agent either improves or preserves its deployed checkpoint as noisy stages accumulate, while naive retraining consistently degrades. The largest final gaps appear on knowledge-intensive and reasoning-heavy tasks where smaller models are most fragile to corrupted training signal: TriviaQA (+43.0pp), ARC-Challenge (+22.6pp), and GSM8K/Llama (+20.3pp). In each case, the naive baseline absorbs poisoned examples---false premises, negation flips, label swaps---that actively teach incorrect associations, while the agent identifies and excludes them. Summarization exhibits the same divergence in a different metric regime: on XSum, naive retraining collapses from ROUGE-2 10.9 to 6.5 while the adaptive agent climbs to 43.1. HumanEval and SAMSum show smaller but still substantial gaps of +17.8pp and +24.6 ROUGE-2, respectively. Notably, the deployed adaptive curve remains monotone or flat by construction: whenever a stage-specific retrain regresses, the system retains the previous checkpoint rather than deploying a worse model.

\subsubsection{Controlled Baseline Comparison.}

To isolate the value of the agent's filtering and curriculum construction from the additional messiness of incomplete human review, we conduct a controlled upper-bound experiment on GSM8K (Qwen3-8B) in which every example---including poisoned ones---receives an oracle correction. This setup gives all baselines access to perfect labels, so any remaining performance difference must arise from how the training data is selected and composed rather than from label quality.

\begin{wraptable}{r}{0.5\textwidth}
\vspace{-\intextsep}
\centering
\small
\setlength{\tabcolsep}{4pt}
\caption{\textbf{Controlled held-out evaluation} on 572 test examples (mixed noisy + clean). Accuracy (\%).}
\begin{tabular}{@{}lrrr@{}}
\toprule
\rowcolor{headergray}
Method & Overall & Noisy & Clean \\
\midrule
Base model (no training) & 56.6 & 49.4 & 66.3 \\
Naive Retrain (40\% poison) & 64.4 & 55.1 & 76.9 \\
Naive Retrain (60\% poison) & 64.3 & 56.0 & 75.2 \\
Random Augmentation & 68.5 & 58.6 & 81.7 \\
Clean Only (no poison) & 71.2 & \cellcolor{lightgreen}\textbf{62.6} & 82.5 \\
\textbf{Pioneer Agent (filtered)} & \cellcolor{lightgreen}\textbf{72.2} & 61.7 & \cellcolor{lightgreen}\textbf{86.2} \\
\bottomrule
\end{tabular}
\label{tab:adaptft-heldout}
\vspace{-1em}
\end{wraptable}

 As shown in Table~\ref{tab:adaptft-heldout}, Pioneer Agent achieves 72.2\% overall accuracy, outperforming the clean-only baseline (71.2\%), random augmentation (68.5\%), and both naive retrain variants (64.3--64.4\%). The gap is most pronounced on the clean regression split, where the agent scores 86.2\% compared to 82.5\% for clean-only and 76.9\% for naive retraining---indicating that the agent's curriculum not only handles noisy inputs better but also preserves previously correct behavior more effectively. That the agent surpasses even the clean-only baseline, which never sees any poisoned data, suggests that selective exposure to fixable noise provides a form of robustness training that pure clean data cannot supply.
 

\subsubsection{Stage-Based Deployment Simulation.}

\begin{wraptable}{r}{0.48\textwidth}
\vspace{-1.2em}
\centering
\small
\setlength{\tabcolsep}{4pt}
\caption{\textbf{GSM8K (Qwen3-8B) stage checkpoints.} Accuracy (\%). Pioneer Agent improves monotonically; naive retraining degrades sharply on the first noisy stage.}
\begin{tabular}{@{}lrrrr@{}}
\toprule
\rowcolor{headergray}
Checkpoint & Naive & Agent & $\Delta$ & Poison \\
\midrule
C1: Clean Base & 71.6 & 75.9 & +4.3 & 0\% \\
C2: +Stage 1   & 62.6 & 77.6 & +15.0 & 15\% \\
C3: +Stage 2   & 63.3 & 79.1 & +15.8 & 25\% \\
C4: +Stage 3   & 64.7 & 81.2 & +16.5 & 40\% \\
\bottomrule
\end{tabular}

\label{tab:adaptft-gsm8k-stages}
\vspace{-1em}
\end{wraptable}

We present a detailed view of the GSM8K (Qwen3-8B) scenario to illustrate the stage-based evaluation under realistic conditions (Table~\ref{tab:adaptft-gsm8k-stages}). Each deployment stage introduces roughly 440 new inference logs with a progressively higher poison rate. The naive baseline retrains once per stage on all incoming logs without filtering---approximately 30\% of noisy failures go unreviewed, and among reviewed poisoned examples roughly half receive incorrect corrections. Pioneer Agent, by contrast, filters poisoned examples, constructs contrastive pairs from fixable noise, and iterates up to twice per stage with rollback.
Despite starting from the same clean checkpoint, the trajectories diverge immediately. On the clean base checkpoint (C1), the gap is a modest 4.3pp, reflecting slightly better data curation even without noise. Once the first noisy stage arrives at 15\% poison rate, naive retraining degrades sharply to 62.6\%---a 9pp drop from its own baseline---while the agent improves to 77.6\%, widening the gap to 15.0pp. Subsequent stages continue this pattern: the naive baseline hovers around 63--65\% as incorrect signal accumulates, while Pioneer Agent climbs to 81.2\%. The final 16.5pp gap reflects not a single decisive intervention but the compounding effect of correct filtering decisions across all three stages.

\begin{table}
\centering
\small
\setlength{\tabcolsep}{6pt}
\renewcommand{\arraystretch}{1.2}

\begin{tabular}{@{}l c c p{8cm}@{}}
\toprule
\textbf{Perturbation Type} & \textbf{Count} & \textbf{Fixable} & \textbf{Agent Action} \\
\midrule

\rowcolor{gray!8}
\multicolumn{4}{@{}l}{\textit{Linguistic / structural noise}} \\
Typo / misspelling & 74 & Yes & Uses contrastive pairs to improve robustness to surface noise \\
Syntactic / grammatical error & 50 & Yes & Uses contrastive pairs to enforce correct interpretation \\
Preamble injection & 69 & Partial & Trains on clean extraction of relevant content \\
Truncation & 28 & Partial & Includes shorter contexts to improve robustness \\
Number distractor & $\sim$50 & Yes & Uses contrastive pairs to resolve numerical confusion \\

\addlinespace[3pt]

\rowcolor{red!4}
\multicolumn{4}{@{}l}{\textit{Semantic poison}} \\
Number swap & $\sim$70 & \textbf{No} & \textbf{Excluded} (induces incorrect arithmetic relationships) \\
False premise & $\sim$60 & \textbf{No} & \textbf{Excluded} (invalidates problem semantics) \\
Negation flip & $\sim$40 & \textbf{No} & \textbf{Excluded} (contradicts the correct answer) \\

\addlinespace[3pt]

\rowcolor{orange!6}
\multicolumn{4}{@{}l}{\textit{Adversarial inputs}} \\
Prompt injection & 49 & \textbf{No} & \textbf{Excluded} (adversarial manipulation) \\
Jailbreak attempt & 49 & \textbf{No} & \textbf{Excluded} (adversarial manipulation) \\

\bottomrule
\end{tabular}

\caption{\textbf{Failure coverage on the GSM8K/Qwen stage benchmark.} The agent separates fixable noise (linguistic and structural) from semantic poison and adversarial inputs. Fixable cases are incorporated into training via targeted supervision, while harmful examples are excluded to prevent learning incorrect behavior. Naive retraining, in contrast, trains on all types indiscriminately.}
\end{table}

\subsubsection{Qualitative analysis.} The mechanism underlying both gaps is \emph{semantic triage}: the agent reasons about whether each failure example is safe to learn from before including it in training. Consider a poisoned math problem where ``bought'' has been changed to ``didn't buy,'' yet the ground-truth answer still assumes the purchase occurred. Naive retraining ingests this example verbatim, reinforcing a contradictory input--label association. The agent detects that the negation alters the problem semantics and excludes the example, retaining only cases where noise is surface-level (e.g., typos, distractors) and the answer remains valid. This per-example filtering is what separates the two trajectories in Tables~\ref{tab:adaptft-heldout} and~\ref{tab:adaptft-gsm8k-stages}, and Figure~\ref{fig:adaptft-stage-trends} confirms that the pattern holds across all seven AdaptFT-Bench scenarios.

\subsection{Deployment Case Studies}

The production mode results above validate the agent under synthetic noise at controlled poison rates. To confirm that the same pipeline transfers to authentic deployment conditions, we evaluate on two additional tasks---CLINC150 (intent classification) and CoNLL-2003 (named-entity recognition)---where the Pioneer Agent platform serves public benchmark data, collects model predictions, and scores them automatically (exact-match for CLINC150, entity-F1 for CoNLL-2003). The production-mode agent receives these judged inference logs directly, with no synthetic perturbation applied.

\subsubsection{Intent Classification (CLINC150)}
\label{sec:clinc150}

We evaluate Pioneer Agent on a production-style deployment of CLINC150 intent classification, focusing on its ability to diagnose failure modes, construct targeted training data, and improve performance under strict regression constraints. A \texttt{gliner2-base-v1} model serves 30-class intent classification over a CLINC150 subset comprising 2,999 production inferences. Under exact-match evaluation, the model passes 84.9\% of inferences (2,546/2,999), leaving 453 failures for the agent to address.

\begin{figure}
\centering
\includegraphics[width=0.98\linewidth]{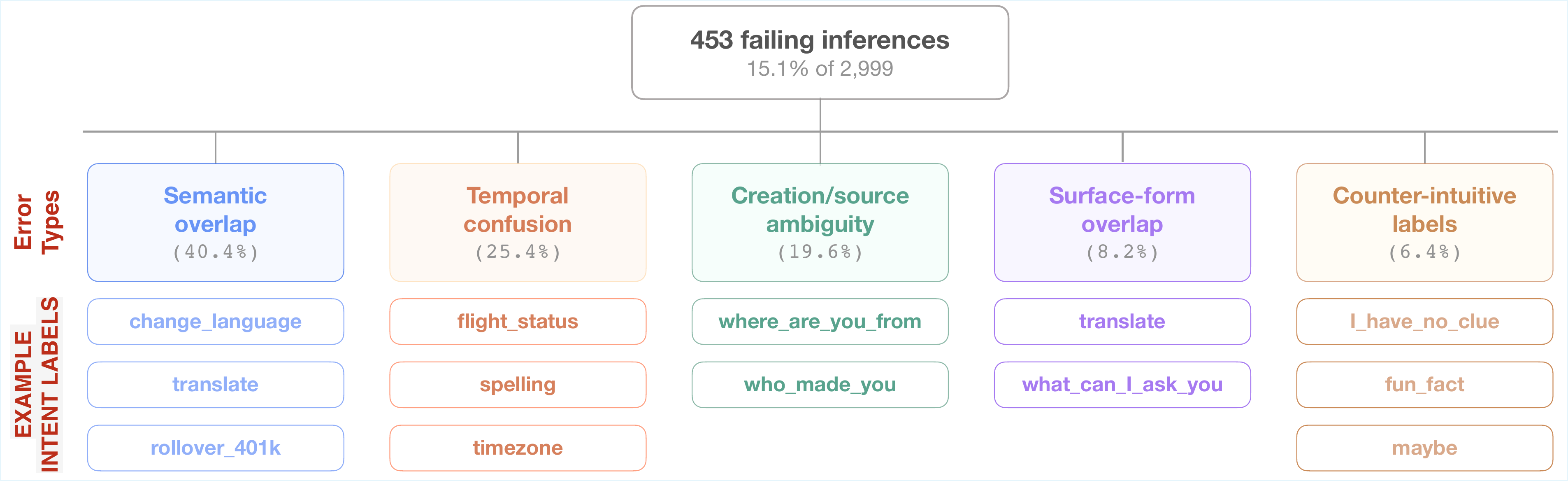}
\caption{\textbf{Failure taxonomy constructed by Pioneer Agent for the CLINC150 intent classification deployment.} The agent clusters 453 production failures into actionable categories by querying the inference database, identifying dominant confusion pairs (e.g., \texttt{flight\_status}$\to$\texttt{time}, \texttt{change\_language}$\leftrightarrow$\texttt{translate}), and diagnosing root causes: semantically overlapping intents, counterintuitive label semantics, and surface-form similarity across decision boundaries. This taxonomy directly informs the composition of the corrective training curriculum.}
\label{fig:failure-taxonomy}
\end{figure}

\textbf{(1)} \textit{Failure diagnosis:}
Rather than treating all 453 failures uniformly, the agent queries the production inference database and constructs the structured failure taxonomy shown in Figure~\ref{fig:failure-taxonomy}. This analysis reveals three distinct root causes. The dominant source is \emph{semantically overlapping intents}: \texttt{change\_language} and \texttt{translate} exhibit fully bidirectional confusion (25 failures each), while \texttt{rollover\_401k} is systematically misclassified as \texttt{transfer} in 23 of 26 cases. A second category stems from \emph{counterintuitive label semantics}---inputs such as ``I have no clue'' belong to the \texttt{maybe} intent, yet the model interprets them as knowledge-seeking and fires \texttt{fun\_fact} (17 of 29 failures), indicating that pragmatic meaning diverges from surface-level cues. The third category involves \emph{surface-form similarity across intent boundaries}: both \texttt{flight\_status} and \texttt{replacement\_card\_duration} queries begin with ``how long,'' causing the model to collapse them into the generic \texttt{time} intent (39/40 and 23/26 failures, respectively). The most failure-prone label overall is \texttt{change\_user\_name} (58 failures), of which 39 are misrouted to the catch-all \texttt{what\_can\_i\_ask\_you}---a pattern that suggests the model lacks sufficient training signal for this label entirely. This diagnostic step shapes every downstream decision: by identifying \emph{which} confusion pairs dominate and \emph{why}, the agent can construct a curriculum that targets the actual failure structure rather than sampling corrections blindly.

\begin{wraptable}{r}{0.6\textwidth}
\centering
\small
\setlength{\tabcolsep}{6pt}
\renewcommand{\arraystretch}{1.2}
\vspace{-1em}
\caption{\textbf{CLINC150 deployment results after one improvement cycle.} The agent fixes nearly all failures while preserving previously correct behavior.}
\begin{tabular}{@{}lccc@{}}
\toprule
\rowcolor{headergray}
\textbf{Metric} & \textbf{Before} & \textbf{After} & $\Delta$ \\
\midrule
Failures fixed & 453 & \textbf{450 (99.3\%)} & 3 remaining \\
Adjusted (excl.\ 1 label error) & 452 & \textbf{450 (99.6\%)} & 2 remaining \\
Regression on passing set & 198 & 197 (99.5\%) & 1 regression \\
\bottomrule
\end{tabular}
\label{tab:clinc150-results}
\end{wraptable}

\textbf{(2)} \textit{Curriculum synthesis and training:}
Guided by the taxonomy above, the agent synthesizes a targeted training curriculum of 1,254 examples. The dataset combines corrected failure cases with systematically constructed hard negatives for each confusion pair---for instance, \texttt{change\_language} queries phrased to resemble \texttt{translate}, and vice versa. In addition, contrastive pairs explicitly teach fine-grained distinctions at decision boundaries, ensuring that semantically similar intents are separated during training. To select the best adaptation strategy, the agent trains two candidate models in parallel---full fine-tuning and LoRA---on the same curriculum. Full fine-tuning achieves near-perfect performance on the first attempt, reaching 99.3\% accuracy on the failure set (450/453) while preserving 99.5\% of previously correct predictions (197/198). LoRA attains 99.3\% on a 150-sample fast-evaluation slice but is not promoted further due to slightly lower stability on the full evaluation set. The agent therefore selects full fine-tuning as the deployment candidate (V1).

\textbf{(3)} \textit{Residual error analysis:}
The three remaining errors fall into distinct categories, each carrying a different implication. One corresponds to a confirmed labeling error in the benchmark---``tell me about today weather'' is annotated as \texttt{replacement\_card\_duration}---which represents an upstream data-quality issue that no amount of training can resolve. The second is a genuine edge case where a \texttt{pto\_request} is phrased as a general inquiry, placing it outside the distribution the curriculum covers. The third reflects a borderline ambiguity between \texttt{what\_can\_i\_ask\_you} and \texttt{definition}, arising from intrinsic overlap in the label space rather than a systematic model shortcoming. By distinguishing these three failure types, the agent avoids the trap of treating every residual error as trainable.

\textbf{(4)} \textit{Refinement attempt and rollback:}
To address the two legitimate remaining failures, the agent introduces 17 additional targeted examples and retrains the model (V2). However, this refinement produces a performance drop to 98.5\%, indicating that the additional data introduces interference with previously learned decision boundaries---a characteristic risk in the high-accuracy regime where the curriculum already covers the vast majority of failure modes. Recognizing this regression, the agent reverts to V1 in accordance with its rollback policy (Section~\ref{sec:iteration-policy}), which forbids dataset expansion to fix residual errors once accuracy exceeds 99\%. This decision demonstrates that effective production adaptation requires not only the ability to learn, but also the discipline to stop.

Table~\ref{tab:clinc150-results} summarizes the final deployment outcome. This case study highlights three properties of the production improvement loop. First, performance gains derive primarily from data quality rather than optimization: a single well-constructed curriculum achieves 99.3\% accuracy without iterative hyperparameter search, confirming that the diagnostic and curriculum-synthesis stages---not the training procedure---carry the explanatory weight. Second, disciplined rollback proves essential in the high-accuracy regime, where additional data can degrade performance by disrupting established decision boundaries. Third, the agent distinguishes model failures from upstream data issues, correctly identifying a mislabeled example as unfixable and excluding it from the training objective. Together, these behaviors show that closing the last few percentage points in production requires not just learning capacity but selective restraint.

\begin{figure}
    \centering
    \includegraphics[width=1.\linewidth]{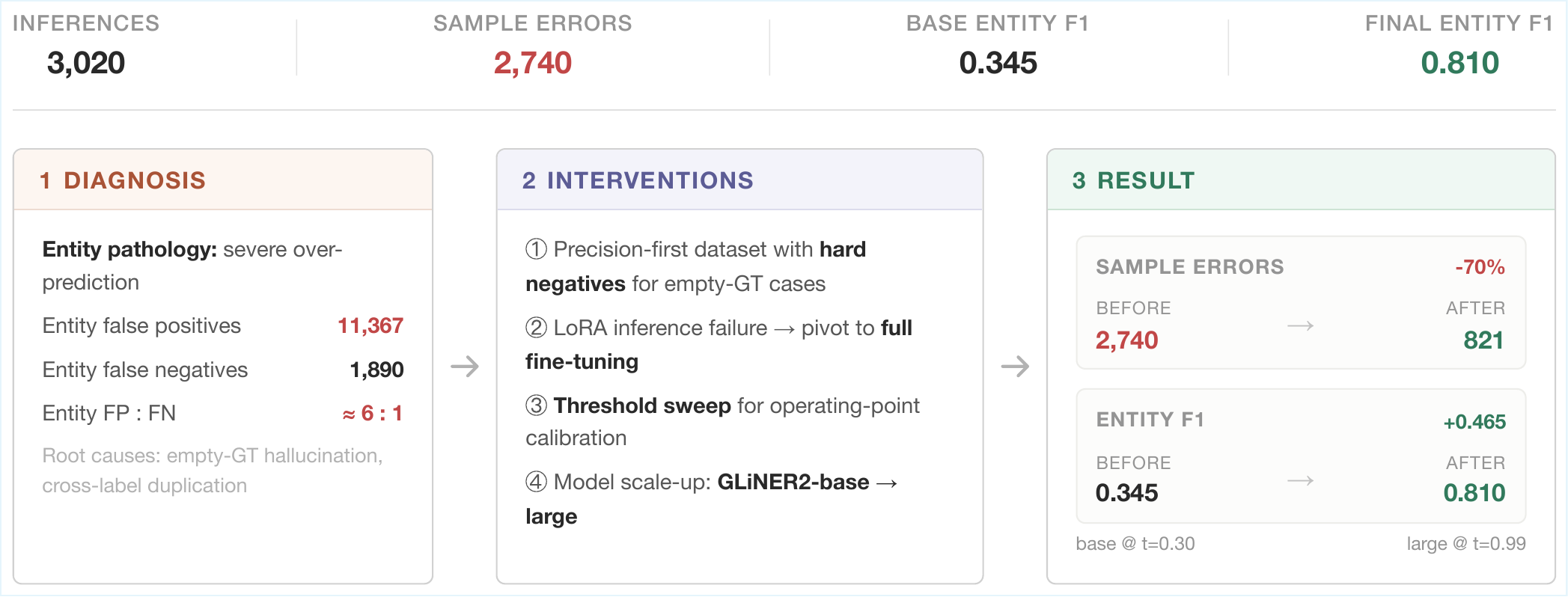}
    \caption{\textbf{Agent-guided recovery on CoNLL-2003 NER.} Starting from a severe precision-collapse regime dominated by false positives, the agent diagnoses the error pathology, builds a precision-first curriculum with hard negatives, pivots from failed LoRA inference to full fine-tuning, and calibrates the deployment threshold. These interventions reduce sample errors from 2,740 to 821 and improve Entity~F1 from 0.345 to 0.810.}
    \label{fig:conll-agent}
\end{figure}

\subsubsection{Named Entity Recognition (CoNLL-2003)}
\label{sec:conll-ner}

We evaluate the agent on a production-style named entity recognition (NER) deployment derived from CoNLL-2003, comprising 3,020 judged inferences. In contrast to the intent classification setting (Section~\ref{sec:clinc150}), errors are not dominated by missed predictions but by systematic over-prediction. A \texttt{gliner2-base-v1} model achieves a low baseline Entity~F1 of 0.345, with only 9.3\% of samples passing. Failures are driven by 11,367 false positives versus 1,890 false negatives—a 6:1 FP\,:\,FN ratio—indicating that the model fires excessively even when no entity is present. After one improvement cycle, the agent raises Entity~F1 to 0.810 and resolves 70\% of the failure slice (Figure~\ref{fig:conll-agent}), combining data curation, training-method pivoting, and threshold tuning into a single adaptive trajectory.

\textbf{(1)} \textit{Diagnosis and precision-oriented curriculum:}
The agent opens with six parallel SQL diagnostics against the inference database and identifies two dominant error modes: \emph{empty-ground-truth hallucination}, where the model predicts entities for inputs that contain none, and \emph{cross-label duplication}, where a single span receives multiple overlapping entity types. Both pathologies inflate the false-positive count, confirming that the core problem is precision rather than recall. Guided by this asymmetry, the agent builds progressively more precision-oriented datasets anchored in large numbers of hard negatives---entity-free passages that penalize hallucination and near-miss spans that penalize label duplication---rather than the balanced positive/negative mix a recall-oriented curriculum would require.

\textbf{(2)} \textit{Training-method recovery and threshold tuning:}
The initial LoRA training run fails at inference with repeated 500 errors, an infrastructure-level failure that would terminate a fixed pipeline. Instead, the agent pivots to full fine-tuning on the same dataset and recovers the training trajectory without manual intervention. Beyond the training method, the agent treats the GLiNER2 entity-score threshold as a searchable parameter: each predicted entity is assigned a probability score, and the default decision threshold is typically 0.5. It therefore sweeps threshold values and compares model capacities between \texttt{gliner2-base-v1} and \texttt{gliner2-large-v1}. The winning configuration pairs GLiNER2-large with a threshold of $t = 0.99$---a highly aggressive precision setting that the agent discovers is necessary precisely because the dominant failure mode is over-prediction. In contrast to CLINC150, where the optimal model emerged from a single training run, this scenario requires the agent to jointly optimize along data, architecture, and inference-time axes.

\paragraph{Takeaways:} This case study surfaces three agent capabilities absent from the CLINC150 scenario. First, the agent distinguishes precision failures from recall failures and reshapes the curriculum accordingly, rather than treating all errors as symmetric. Second, it recovers from infrastructure-level failures by switching training methods mid-trajectory instead of abandoning the run---a form of robustness that goes beyond data curation. Third, it performs deployment threshold tuning, discovering that threshold $t = 0.99$ is part of the winning policy, not merely an evaluation detail. The full trajectory is analyzed in Section~\ref{sec:conll-ner-traj}.

\section{Analysis}



\subsection{Emergent Training Strategies}
 
A central question in autonomous fine-tuning is whether strong performance arises from explicitly programmed heuristics or from the agent’s ability to discover effective strategies through interaction with the task. In particular, it is unclear whether an agent operating purely from outcome-based feedback can recover the nuanced training practices that human practitioners typically design through extensive experimentation and domain knowledge. We therefore examine whether such strategies can emerge purely from feedback-driven optimization.

Empirically, we find that they do. Across all experiments (both \textit{cold-start} and \textit{production}), no run was explicitly instructed which training strategies to use. Yet the agent consistently recovers techniques that are individually well-established in the literature---including chain-of-thought supervision, adaptive epoch scheduling, and quality-focused data curation---and composes them into effective pipelines guided solely by downstream performance feedback. We organize the findings into three groups, each of which is developed in the following subsections.

\paragraph{(1) Supervision Format and Prompting.}
Two of the agent's most impactful interventions target \emph{how} examples are presented rather than \emph{which} examples are used.

\begin{figure}
\centering
\includegraphics[width=0.98\linewidth]{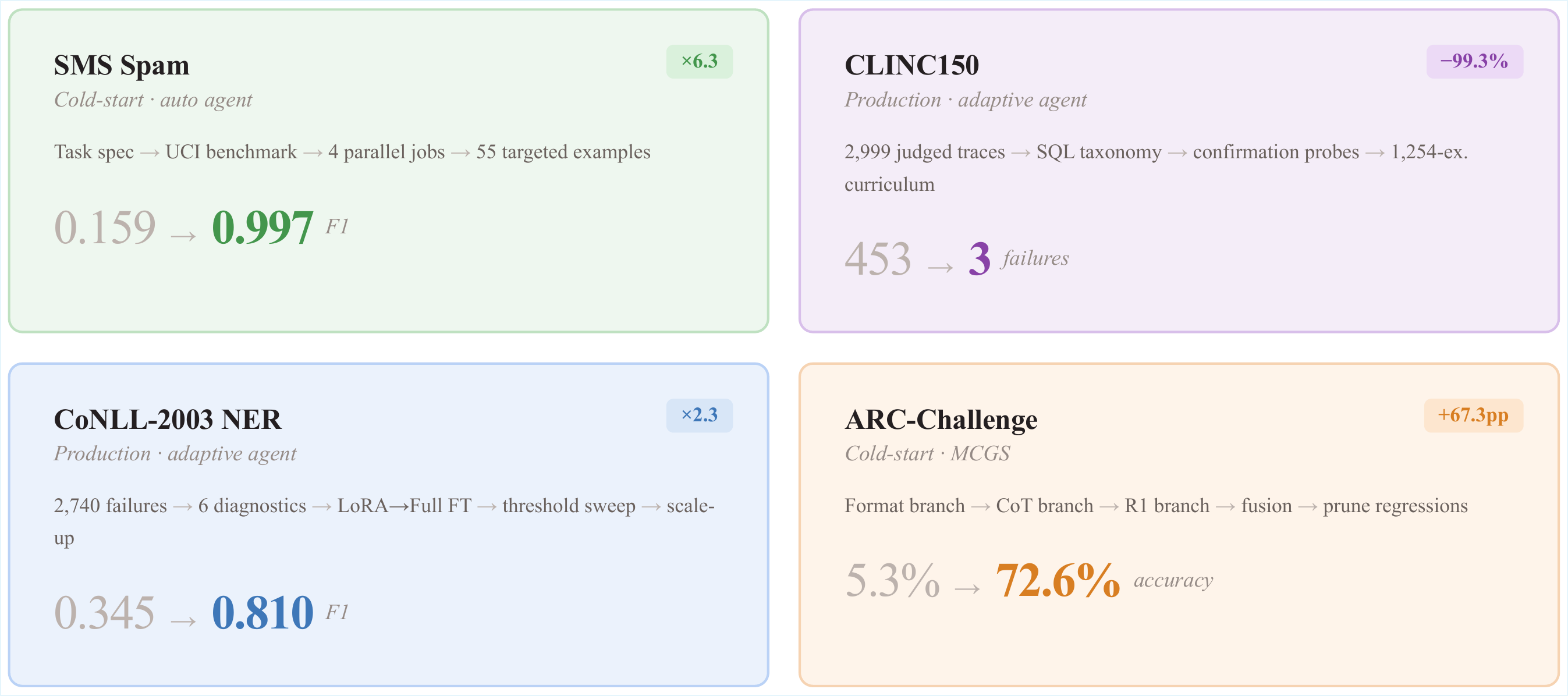}
\caption{\textbf{Four representative Pioneer Agent trajectories spanning cold-start and production settings.} Each panel compresses a full run into its diagnostic signal, the intervention sequence the agent pursued, and the final validated outcome. Despite differences in task family, tooling, and search strategy, all four trajectories follow the same closed-loop pattern: diagnose failures, construct targeted supervision, test competing interventions, and retain only updates that survive evaluation.}
\label{fig:trajectory-motifs}
\end{figure}

For reasoning-intensive benchmarks, the agent consistently favored chain-of-thought over direct-answer supervision. On ARC-Challenge, this transition increased accuracy from 48\% to 69\%, the single largest gain across all benchmarks. The agent further discovered that DeepSeek-R1 reasoning traces outperformed GPT-4.1 traces, reaching 72.6\% after fusion (Figure \ref{fig:trajectory-motifs}). This behavior emerged specifically for reasoning tasks, indicating that the agent correctly deduced the appropriate supervision format from observed failure modes.

Separately, the agent treated the system prompt as a tunable configuration component. On SAMSum, replacing a generic prompt (``summarize this dialogue'') with a constrained specification (``1--3 sentences, main outcome, third person, brief'') produced the single largest improvement of any intervention. The agent attributed output variability to prompt under-specification rather than a data problem---a distinction that avoids wasted iterations on data augmentation.

\paragraph{(2) Optimization Dynamics.}
In addition to supervision format, the agent adapts \emph{how} models are trained by tuning optimization dynamics to task-specific characteristics through direct comparison.

First, training duration varies markedly across tasks. The agent selects 1 epoch for XSum (where summarization overfits rapidly), 2 for GSM8K (combined with larger batch sizes to stabilize gradients), 5 for ARC-Challenge, and 8 for SAMSum (a low-data regime). To make these decisions, it jointly monitors validation and training performance to detect divergence or premature convergence, treating epoch count as the primary corrective lever.

Similarly, learning rate selection follows a task-dependent pattern. Knowledge-intensive tasks (TriviaQA, ARC-Challenge) converge on lower rates ($1\mathrm{e}{-4}$) that better preserve pretrained factual knowledge, whereas format-learning tasks (HumanEval, SAMSum) tolerate higher rates ($2\mathrm{e}{-4}$). For example, on TriviaQA, $\text{LR}=2\mathrm{e}{-4}$ yields 46.0\% compared to 48.6\% at $1\mathrm{e}{-4}$.

\paragraph{(3) Data Curation Decisions.}
Finally, the agent adapts \emph{what} data to train on, making selective inclusion and exclusion decisions that directly shape the training signal.

First, the agent consistently favors smaller, higher-quality datasets over larger but noisier alternatives. On HumanEval, 173 curated examples outperform 348 (100\% vs.\ 94.5\%). Similarly, on SAMSum, 500 agent-selected examples outperform more than 2,000 randomly sampled instances. On ARC-Challenge, the agent identifies and discards a noisy external data mix, improving performance from 69.6\% to 72.6\%. These results indicate that data quality, rather than scale alone, is the dominant factor.

Moreover, the agent recognizes that external model outputs can dilute the training signal. On HumanEval, incorporating GPT-4.1-generated solutions reduces performance relative to using the model's own correct outputs (96.9\%$\to$94.5\%). In response, the agent reverts to self-generated labels, reasoning that stylistically mismatched external outputs conflict with the base model's learned distribution.

Taken together, these behaviors show that the agent does not treat data aggregation as monotonic, but instead actively filters and aligns the training distribution with the target model.

\subsection{Trajectory Motifs}

To validate that the architectural mechanisms from Section~\ref{sec:pioneer-agent} operate as intended, we analyze behavioral patterns that recur consistently across runs. Figure~\ref{fig:trajectory-motifs} provides an overview of four representative trajectories spanning both agent modes, three task families, and two search strategies.

We structure the analysis around four recurring motifs, each corresponding to a specific design mechanism. Full step-by-step narratives are provided in Appendix~\ref{app:cold-start-trajectories} (SMS Spam, ARC-Challenge), while concise summaries of the resulting artifacts appear in Appendix~B (CLINC150, CoNLL-2003).


\subsubsection{Broad-to-Surgical Exploration}

The iteration policy (Section~\ref{sec:iteration-policy}) predicts that the agent should begin with broad, parallel exploration and narrow to surgical augmentation as accuracy improves. This pattern appears consistently. On SMS Spam, four parallel initial jobs establish a strong baseline (F1 0.9834), after which three rounds adding only 55 total examples drive F1 to 0.9967---each round targeting a specific precision or recall failure identified by error analysis. On CLINC150, the agent invests its initial effort in failure-taxonomy construction (clustering 453 failures into five dominant groups with root-cause attribution), then builds a single 1,254-example curriculum that achieves 99.3\% on the first training run. In both cases, the diagnosis phase consumes more agent reasoning turns than the training phase itself.

\subsubsection{Rollback as a Control Mechanism}

The rollback-first principle (Section~\ref{sec:iteration-policy}, Rule~4) activates across multiple benchmarks, preventing the cascading regressions that typically plague iterative fine-tuning. On CLINC150, adding 17 targeted examples for two remaining failures regressed accuracy from 99.3\% to 98.5\%; the system rolled back rather than attempting to compensate with further augmentation. On ARC-Challenge, iterations 7--11 all attempted to improve on the fusion peak of 72.6\% through more data, higher LoRA rank, or lower learning rates---all were pruned. On SMS Spam, successive iterations alternated between recall repair (adding spam examples) and precision repair (adding hard ham examples), with the regression gate catching each overcorrection. The pattern demonstrates that restraint at high accuracy is as important as exploration at low accuracy, and that the system's conservative regression threshold ($\epsilon = 2$) prevents overfitting to the most recent failure set.

\subsubsection{Error-Mode-Driven Search Pivots}
\label{sec:conll-ner-traj}

The CoNLL-2003 trajectory illustrates how failure diagnosis can fundamentally reshape the search strategy. The agent's first action---six parallel SQL queries against the inference database---reveals a 6:1 FP:FN ratio (11,367 false positives vs.\ 1,890 false negatives), transforming the objective from ``improve entity recall'' to ``suppress entity hallucination.'' This finding drives three interventions, in order of priority: (1)~precision-first data construction with large proportions of empty-ground-truth hard negatives, (2)~threshold calibration via a sweep to 0.99, and (3)~model capacity scaling from GLiNER2-base to GLiNER2-large. The ordering reflects a design principle: exhaust cheaper interventions (data, calibration) before testing whether model capacity is the bottleneck. This trajectory also demonstrates infrastructure recovery: when LoRA inference failed with repeated 500 errors, the agent diagnosed the issue as a platform-level adapter-loading problem and pivoted to full fine-tuning rather than terminating the run.

\subsubsection{Cross-Branch Fusion in Graph Search}

The ARC-Challenge MCGS trajectory (Figure~\ref{fig:mcgs-arc}) demonstrates how graph-structured search enables strategy recombination that linear ablation cannot achieve. Branch~A establishes that format learning is necessary but insufficient (5.3\%$\to$61.3\%). Branch~B discovers that chain-of-thought supervision is the largest observed lever (+21pp). Branch~C identifies DeepSeek-R1 as a marginally better teacher than GPT-4.1 (68.9\% vs.\ 68.6\%). The winning configuration (72.6\%) fuses the R1 reasoning branch with the validation-data expansion strategy from another branch---a recombination of independently discovered strategies. Post-fusion pruning confirms convergence: five subsequent expansions with varied hyperparameters all regress to the 68.7--72.4\% range, validating the progressive exploit shift described in Section~\ref{sec:search-space}.

\section{Discussion}

\subsection{Cost Analysis}

Each Pioneer Agent run incurs two primary categories of cost: LLM orchestration and training infrastructure. 

For orchestration, we estimate approximately 4M tokens per 12-hour run, split roughly 75/25 between input and output. This skew reflects the system's read-heavy interaction pattern: at each reasoning step, the agent ingests evaluation results, trace data, and prior context, while producing relatively short tool calls and analyses. Under Claude Sonnet 4.6 pricing (\$3/1M input tokens, \$15/1M output tokens), this corresponds to approximately \$24 in API fees for a full 12-hour run. Shorter runs scale proportionally; for example, the 1-hour CLINC150 run incurred under \$2 in orchestration cost.

Training and inference are executed on Tinker infrastructure, while the agent's Modal sandbox (CPU-only, 16GB memory) is billed at \$0.50/hour. GPU costs for LoRA fine-tuning and batch inference are included within the Tinker platform fee and vary with model size and dataset volume. In practice, encoder training (GLiNER2) typically completes within 2--5 minutes per run, whereas decoder training (Qwen3-8B, Llama~3.2-3B) requires 10--30 minutes.

Overall cost is driven primarily by the number of iterations and the per-iteration training footprint. This interaction explains the variance observed across benchmarks. For instance, XSum requires 13 iterations over 3,000-example datasets, accumulating \$35 in training costs alone, while HumanEval converges in 4 iterations for under \$5.

Table~\ref{tab:cost-analysis} summarizes representative end-to-end costs across both production improvement and cold-start settings. Total cost ranges from \$3.48 to \$54.98, depending on task complexity, iteration count, and branching behavior. Notably, even the most expensive runs remain substantially below the cost of equivalent manual effort, where a human ML engineer (e.g., at \$150/hour) would incur significantly higher expense.

\begin{table*}
\centering
\begin{tabularx}{\linewidth}{@{} X c rr r @{}}
\toprule
 & & \multicolumn{2}{c}{Cost Breakdown} & \\
\cmidrule(lr){3-4}
Run & Hours & Claude API & Train + Sandbox & \textbf{Total} \\
\midrule
\multicolumn{5}{@{}l}{\textit{Production improvement (AdaptFT-Bench)}} \\[3pt]
GSM8K (3-stage, run A) & 11.5 & \$22.98 & \$10.00 & \$32.98 \\
GSM8K (3-stage, run B) & 8.3 & \$16.62 & \$9.00 & \$25.62 \\
CLINC150 (2 iterations) & 1.0 & \$1.98 & \$1.50 & \$3.48 \\
\midrule
\multicolumn{5}{@{}l}{\textit{Cold-start}} \\[3pt]
ARC-Challenge & 12.0 & \$24.00 & \$12.00 & \$36.00 \\
HumanEval & 4.0 & \$7.98 & \$4.50 & \$12.48 \\
XSum & 10.0 & \$19.98 & \$35.00 & \$54.98 \\
SAMSum & 8.0 & \$16.02 & \$12.00 & \$28.02 \\
\bottomrule
\end{tabularx}
\caption{\textbf{Estimated cost per benchmark run.} Total cost ranges from \$3.48 to \$54.98. The two GSM8K rows are representative end-to-end three-stage adaptive runs; total cost varies with branching, diagnosis, and rollback overhead. A human ML engineer at \$150/hour would cost substantially more for equivalent work.}
\label{tab:cost-analysis}
\end{table*}

\subsection{Failure Modes}

Across all experiments, several recurring failure modes emerge, reflecting both intrinsic model limitations and constraints of the fine-tuning process.

\textbf{Model capacity limitations.}
Certain errors stem from fundamental limitations of the base model rather than deficiencies in the training pipeline. On TriviaQA \citep{joshi2017triviaqa} (Llama~3.2-3B), the agent correctly diagnosed that ``89\% of failures are genuine factual knowledge gaps'' in the closed-book setting. Because these errors reflect missing parametric knowledge, additional fine-tuning data cannot resolve them at this scale. The agent converged at 48.6\% and appropriately terminated further iterations.

\textbf{High-baseline plateaus.}
When the base model already operates near saturation (e.g., Qwen3-8B on ARC at 91.7\%), fine-tuning becomes unstable and can initially degrade performance. On ARC, the first training run reduced accuracy from 91.7\% to 87.8\% before subsequent iterations recovered and improved it to 93.3\%. This highlights the difficulty of making incremental gains in high-performance regimes, where small distribution shifts can outweigh benefits.

\textbf{Adversarial inputs unfixable by training.}
A subset of failures arises from inputs that violate task assumptions, such as prompt injection, jailbreak attempts, or false premises. These cases cannot be corrected through supervised fine-tuning. On GSM8K \citep{cobbe2021training}, the agent identified 17 out of 100 such examples and excluded them from training, preventing contamination of the learning signal.

\textbf{Overfitting in later iterations.}
Performance typically improves over the first 2--4 iterations before regressing as the model begins to overfit to increasingly narrow data slices. While the agent's rollback mechanism mitigates permanent degradation, it does not eliminate wasted computation from unproductive runs, indicating a limit to iterative refinement without stronger regularization or stopping criteria.

\textbf{Instruction sensitivity.}
Agent performance is sensitive to task specification. In one adaptive fine-tuning run, accuracy dropped to 11\% because evaluation was performed without the required system prompt despite explicit instructions. Providing clearer and more explicit task descriptions improved overall efficiency by 2.3$\times$, underscoring the importance of precise specification in autonomous settings.

\textbf{Format-specific benchmarks.}
Finally, some benchmarks impose strict output formatting constraints (e.g., regex-validated answers) that do not generalize from standard training data. The agent correctly identifies these cases as requiring format-specific supervision rather than general capability improvement, adjusting its strategy accordingly.

\subsection{Limitations}

While the results demonstrate strong performance across a range of settings, several limitations remain.

\textbf{Model families.}
The current system supports a limited set of model architectures: GLiNER2 (encoder) \citep{zaratiana-etal-2025-gliner2} and Qwen/Llama (decoder) \citep{touvron2023llama,touvron2023llama2}. Extending to other modalities, such as vision models or embedding models, as well as additional architectures, would require new tooling and integration work.

\textbf{Training methods.}
Support for training paradigms is similarly constrained. Decoder models rely exclusively on LoRA fine-tuning \citep{hu2021lora}, while encoder models (GLiNER2) support both full fine-tuning and LoRA. More advanced alignment methods, such as RLHF \citep{ouyang2022instructgpt} or DPO \citep{rafailov2023dpo}, are not yet incorporated, limiting applicability in settings that require preference optimization.

\textbf{Evaluation dependency.}
The production improvement setting depends on access to reliable evaluation signals, either through LLM-as-judge \citep{zheng2023judging,jiang2024llmjudge,li2024llmjudges,chen2024judgebias,ho2025llmjudgeqa} or human annotations on inference logs. As a result, deployments without systematic quality monitoring cannot directly benefit from the agent.

\textbf{Single-platform evaluation recipe.}
Empirical validation is tied to a single evaluation stack: real-world case studies are drawn from Pioneer Agent deployments, and synthetic scenarios are generated using our perturbation pipeline. The extent to which these results generalize across different platforms, logging infrastructures, or data distributions remains an open question.

\textbf{Agent cost.}
Finally, the agent itself incurs non-trivial cost. The orchestrator LLM (Claude Sonnet 4.6) adds fixed overhead in token usage and latency, which can dominate in lightweight settings. For simple fine-tuning tasks—such as small hyperparameter sweeps or clean datasets—this overhead may outweigh the benefits of autonomous reasoning, as a human engineer could achieve similar results more efficiently. The agent’s iterative loop (diagnosis, data construction, training, evaluation) further compounds cost through repeated cycles. However, this trade-off shifts with task complexity. In settings with noisy data or interacting failure modes, the agent’s ability to diagnose errors and build targeted training curricula can offset its overhead. It effectively replaces multiple rounds of manual debugging with a structured search process, making it most cost-effective when naive iteration is costly or unreliable.

\section{Related Work}
\label{sec:related-work}

\subsection{Automated ML, Data-Centric AI, and LM-Program Optimization}

\textbf{Automated machine learning.}
Classical AutoML work studies automatic model selection and hyperparameter optimization over predefined search spaces. Auto-WEKA \citep{thornton2013autoweka} and Auto-sklearn \citep{feurer2015autosklearn} formulate this as combined algorithm selection and hyperparameter optimization, while Hyperband \citep{li2017hyperband} and BOHB \citep{falkner2018bohb} emphasize resource-aware search, strong anytime performance, and fast convergence. These systems automate important parts of model development, but they operate over predefined configuration spaces and do not reason directly about deployment failures or construct corrective supervision.

\textbf{Data-centric AI and supervision construction.}
A related line of work treats the dataset itself as the primary object of optimization. \cite{zha2023datacentric} provide a survey of data-centric AI, organizing methods for data development, maintenance, and debugging. Cleanlab \citep{northcutt2021confident} focuses on identifying label issues and estimating label uncertainty, while DataPerf \citep{mazumder2023dataperf} benchmarks data-centric algorithms. Data programming \citep{ratner2016dataprogramming} and Snorkel \citep{ratner2020snorkel} study programmatic supervision through noisy labeling functions, Data Shapley \citep{ghorbani2019datashapley} studies the value of individual training examples, and active learning \citep{settles2009active} studies strategies for selecting informative unlabeled examples to query for labeling. S3 \citep{wang2023letssynthesizestepstep} is especially relevant because it iteratively synthesizes data by extrapolating the errors of a small model on a small real validation set. 

\textbf{Prompt and LM-program optimization.}
Another line of work optimizes how language models are used rather than how their weights are adapted. OPRO \citep{yang2023opro}, Automatic Prompt Engineer \citep{zhou2023ape}, and Promptbreeder \citep{fernando2023promptbreeder} search over prompts or instructions. DSPy \citep{khattab2023dspy} and MIPROv2 \citep{opsahl-ong2024optimizing} optimize modular LM programs at the level of instructions and demonstrations. TextGrad \citep{yuksekgonul2024textgrad} and Reflexion \citep{shinn2023reflexion} use textual feedback or verbal reflection to improve components of compound AI systems and agent behavior without directly updating the deployed model through a failure-driven retraining loop. GEPA (Genetic-Pareto) \citep{agrawal2026gepareflectivepromptevolution} combines the approaches of search-based prompt optimizers \citep{yang2023opro, zhou2023ape, fernando2023promptbreeder, khattab2023dspy, opsahl-ong2024optimizing} and textual/reflection-based improvers \citep{yuksekgonul2024textgrad, shinn2023reflexion}. GEPA samples trajectories (e.g., reasoning, tool calls, and tool outputs) from an AI system containing one or more LLM prompts and reflects on them in natural language to diagnose problems, propose and test prompt updates, and combine complementary lessons from the Pareto frontier of its own attempts.

Taken together, these lines of work automate model choice, data construction, and LM-program interfaces, but they do not study the full closed-loop adaptation problem considered here: constructing supervision from judged failures, retraining the model itself, and verifying improvement with explicit regression checks.

\subsection{Agentic Search, Autonomous ML Engineering, and Post-Training}

\textbf{Execution-grounded agentic search.}
Recent systems show that LLM-guided search can be effective when coupled with execution feedback. FunSearch \citep{romera-paredes2024funsearch} pairs an LLM with a systematic evaluator to evolve programs for mathematical and algorithmic discovery. AVO \citep{chen2026avo} uses autonomous coding agents as variation operators in evolutionary search and discovers GPU kernels that outperform strong hand-optimized baselines. Both systems, like ours, use real execution as the fitness signal, but they search over code transformations rather than the joint space of training data, hyperparameters, and learning strategies.

\textbf{Autonomous scientific discovery.}
A closely related set of works studies long-horizon scientific workflows. The AI Scientist \citep{lu2024aiscientist} automates idea generation, experimentation, analysis, and paper writing. The AI Scientist-v2 \citep{yamada2025ai} removes reliance on human-authored code templates and uses agentic tree search to conduct end-to-end scientific research. AI co-scientist \citep{gottweis2025towards} studies a multi-agent generate--debate--evolve loop for producing and refining research hypotheses. These systems generate scientific knowledge; our system generates deployment-ready models, and our target problem is narrower: repeated model adaptation under concrete task and regression constraints.

\textbf{Recursive self-improvement.} 
From the Gödel Machine \citep{schmidhuber2006goedelmachinesselfreferentialuniversal} to recent LLM self-improvement loops \citep{huang2022largelanguagemodelsselfimprove, yuan2025selfrewardinglanguagemodels}, studies have been done on systems that iteratively refine their own behavior. Huxley-Gödel Machine \citep{wang2025huxleygodelmachinehumanlevelcoding} applies clade-aware tree search to address the observation that the currently best-performing node in a search tree is not always the most productive parent for generating strong descendants.

\textbf{Autonomous ML and data-science agents.}
Several recent benchmarks and systems study whether agents can perform substantial portions of machine learning or data-science workflows. MLAgentBench \cite{huang2023mlagentbench} evaluates end-to-end machine learning experimentation on research-style tasks. DSBench \cite{jing2024dsbench} broadens this view to realistic data analysis and data modeling tasks. MLE-bench \cite{chan2024mlebench} evaluates agents on Kaggle-style machine learning engineering competitions, AIDE \cite{jiang2025aide} frames ML engineering as tree search in the space of code, MLE-STAR \cite{nam2025mle} combines web retrieval with targeted refinement of components in the ML pipeline, and MLEvolve \cite{du2025automlgen, feng2026internagent} applies MCGS to Kaggle competition solving. We adapt the same MCGS framework for fine-tuning, where nodes represent training pipelines rather than competition submissions, and our production mode adds a dimension absent in these systems: grounding the search in real deployment failures rather than static metrics. Tool-using LLM agents such as ReAct \cite{yao2023reactsynergizingreasoningacting}, Voyager \cite{wang2023voyageropenendedembodiedagent}, and SWE-Agent \cite{yang2024sweagentagentcomputerinterfacesenable} demonstrate that language models can serve as effective controllers over long-horizon tool-use trajectories; OpenHands \cite{wang2024openhands} provides an open platform for building such agents with sandboxed execution. Our system applies this capability to the specific domain of model fine-tuning, where the ``tools'' are training scripts, data generators, and evaluation harnesses. Open fine-tuning frameworks such as LLaMA-Factory \cite{zheng2024llamafactory} and Axolotl \cite{axolotl2024} represent the standard human-operated tooling that Pioneer Agent aims to automate.

\textbf{Autonomous fine-tuning and post-training.} Direct Preference Optimization \cite{rafailov2023dpo}, Reinforcement Learning from Human Feedback \cite{ouyang2022rlhf}, Kahneman-Tversky Optimization \cite{ethayarajh2024kto}, and constitutional AI \cite{bai2022constitutional} are the dominant paradigms for post-training LLMs from feedback signals.  But these methods focus on broad alignment from human preferences instead of our setting of task-specific adaptation from execution failures. Two recent concurrent works are especially close to our setting. FT-Dojo \cite{li2026ft} introduces an interactive environment for end-to-end autonomous LLM fine-tuning across multiple domains. PostTrainBench \cite{rank2026posttrainbench} studies whether agents can autonomously post-train base language models under bounded compute constraints.
These works share our interest in autonomous model training but differ in three respects. First, they focus on \emph{offline} optimization against static benchmarks; Pioneer Agent adds a \emph{production} mode grounded in live inference failures. Second, their search is over training hyperparameters on a fixed dataset; Pioneer Agent searches jointly over data composition, hyperparameters, and learning strategy ($D \times H \times S$; Section~\ref{sec:search-space}). Third, neither system addresses regression testing or failure-driven curriculum synthesis -- the mechanisms needed to improve a model that is already deployed.

\subsection{Failure-Driven Adaptation, Curriculum Construction, and Continual Improvement}

\textbf{LLM-as-a-Judge and automated quality signals.}
Our production loop assumes access to judged failures, either from human review or from an automated judge. Recent work shows that strong LLM judges can provide scalable quality signals for open-ended outputs \citep{zheng2023judging}. Follow-up work trains open judge models such as Prometheus \citep{kim2024prometheus} and JudgeLM \citep{zhu2025judgelm}, and recent surveys study reliability, bias, and propose evaluation methodology for LLM-as-a-Judge systems \citep{gu2024llmajudge}.

\textbf{Curriculum learning and adaptive example selection.}
Curriculum learning \citep{bengio2009curriculum} and self-paced learning \citep{kumar2010selfpaced} study how the order or selection of examples affects learning. Teacher--Student Curriculum Learning \citep{matiisen2019teacher} further automates curriculum selection by choosing subtasks based on learning progress. Pioneer Agent does not treat the training set as fixed, instead it builds a targeted curriculum from observed failures.

\textbf{Continual learning, replay, and model degradation.}
Continual learning studies how to adapt models over time without catastrophic forgetting \citep{parisi2019continual}. Representative approaches include standard replay buffers, episodic-memory methods such as GEM \citep{lopezpaz2017gem}, and parameter-regularization methods such as EWC \citep{kirkpatrick2017ewc}. More recent work has studied continual learning specifically for language models. O-LoRA \citep{wang2023orthogonal} learns new tasks in low-rank subspaces that are constrained to be orthogonal to previously learned subspaces, reducing interference without storing replay data. OSFT \citep{nayak2025sculpting} preserves high-rank parameter subspaces identified by Singular Value Decomposition and constrains new updates to orthogonal low-rank subspaces, improving the trade-off between adaptation and retention. In parallel, concept-drift and model-monitoring work studies how deployed models degrade over time and how such shifts can be detected \citep{gama2014conceptdrift, bayram2022modeldegradation}. Our production setting also requires preserving prior behavior while responding to new failure patterns.

Prior work in this bucket typically studies automated evaluation, curriculum selection, continual adaptation, or drift detection in isolation. Pioneer Agent combines these ingredients into a single closed-loop repair process: it uses judged failures to build a failure taxonomy, confirms weaknesses through targeted probing, synthesizes corrective data, retrains the model, and accepts an update only if it fixes targeted failures without introducing regressions.

\section{Conclusion}

We presented Pioneer Agent, a closed-loop system for autonomous fine-tuning of small language models. The system operates in two modes -- cold-start (from task description) and production improvement (from inference failures) -- and uses agent-guided iterative search---including Monte Carlo Graph Search in selected runs---to navigate the joint space of training data, hyperparameters, and learning strategies.

On eight cold-start benchmarks, the agent achieves gains of +21--84\% over base-model baselines---including format-learning gains for non-instruction-tuned models---with the largest improvements on SMS Spam (+83.8\% F1), ARC-Challenge (+67.3pp), and TriviaQA (+48.6\%). On AdaptFT-Bench, which uses synthetic inference logs with accumulating noise to simulate production drift, Pioneer Agent improves or preserves its deployed checkpoint across seven scenarios while naive retraining degrades, with final gaps of up to 43\%. The same loop improves staged deployment slices from 84.9\% to 99.3\% on CLINC150 and from Entity~F1 0.345 to 0.810 on CoNLL-2003.

Beyond aggregate gains, the agent autonomously selects effective fine-tuning strategies---chain-of-thought training for reasoning, task-specific epoch tuning, quality-over-quantity data curation, and system prompt engineering---based on observed performance feedback rather than explicit instruction.

The full fine-tuning lifecycle from task specification through production deployment can be substantially automated. We release the perturbation pipeline used to build the synthetic inference scenarios at \url{https://github.com/fastino-ai/synthetic-inference-logs} to support further research on autonomous model improvement agents.

\bibliographystyle{plainnat}
\bibliography{references}   

\begin{appendices}

\section{Monte Carlo Graph Search Algorithm}
\label{app:mcgs}

Monte Carlo Graph Search (MCGS) is used to explore the space of training pipelines. The search maintains a graph of evaluated pipelines and iteratively expands promising nodes based on their observed performance. Node selection is guided by a UCT-like score:
\begin{equation}
    \text{UCT}(v_i) = \bar{f}(v_i) + c(t) \sqrt{\frac{\ln N}{n_i}}
    \label{eq:uct}
\end{equation}
where $\bar{f}(v_i)$ is the mean reward of $v_i$'s descendants, $N$ is the total visit count, $n_i$ is the visit count for $v_i$, and $c(t)$ is a time-decaying exploration coefficient. In practice, $c(t)$ is not set to a fixed functional form; the orchestrator LLM adjusts the exploration--exploitation balance heuristically based on the iteration count, the spread of scores in the current graph, and whether recent expansions have yielded improvements. The stagnation threshold (number of consecutive non-improving expansions before triggering fusion or evolution) is similarly determined by the agent rather than a fixed constant. This means the search policy is partially implicit in the LLM's reasoning rather than fully specified by closed-form hyperparameters---a design trade-off that increases flexibility at the cost of exact reproducibility. Of the eight cold-start benchmarks, explicit graph-structured MCGS with branching and fusion was used for ARC-Challenge (Llama 3B); the remaining benchmarks used sequential greedy iteration guided by the same diagnose--modify--evaluate loop without maintaining a formal search graph.

\begin{center}
\begin{minipage}{0.92\textwidth}
\small
\begin{algorithm}[H]
\SetAlgoLined
\SetAlgoSkip{smallskip}
\setlength{\algomargin}{1em}

\KwIn{failure set or task specification $\mathcal{F}$; evaluation function $f(\pi)$ that trains pipeline $\pi$ and returns its validation-set score}
\KwOut{best pipeline $\pi^\star$}

$\pi_0 \gets \textsc{Baseline}(\mathcal{F})$\;
$s_0 \gets f(\pi_0)$\;
$v_0 \gets (\pi_0, s_0)$\tcp*{node = (pipeline, realized score)}
$\mathcal{V} \gets \{v_0\}$, $\mathcal{E} \gets \emptyset$, $\mathcal{G} \gets (\mathcal{V}, \mathcal{E})$\;

\While{\textsc{Continue}$(\mathcal{G}, \mathcal{F})$}{
    \tcp{Select a promising leaf node}
    $v_{\mathrm{parent}} \gets \arg\max_{v \in \textsc{Leaf}(\mathcal{G})} \textsc{UCT}(v)$\;

    \tcp{Expand by proposing a new pipeline from the parent trajectory}
    $\pi_{\mathrm{new}} \gets \textsc{Agent}(v_{\mathrm{parent}}, \mathcal{G}, \mathcal{F})$\;
    $s_{\mathrm{new}} \gets f(\pi_{\mathrm{new}})$\;
    $v_{\mathrm{new}} \gets (\pi_{\mathrm{new}}, s_{\mathrm{new}})$\;

    $\mathcal{V} \gets \mathcal{V} \cup \{v_{\mathrm{new}}\}$\;
    $\mathcal{E} \gets \mathcal{E} \cup \{(v_{\mathrm{parent}}, v_{\mathrm{new}})\}$\;
    $\mathcal{G} \gets (\mathcal{V}, \mathcal{E})$\;

    \tcp{Backpropagate score statistics through ancestors}
    \textsc{Backprop}$(\mathcal{G}, v_{\mathrm{new}})$\;

    \tcp{Recover from stagnation by evolution or fusion}
    \If{\textsc{Stagnant}$(\mathcal{G}, v_{\mathrm{parent}})$}{
        \uIf{\textsc{EvolvePreferred}$(\mathcal{G}, v_{\mathrm{parent}})$}{
            $\pi_{\mathrm{evo}} \gets \textsc{Evolve}(v_{\mathrm{parent}}, \mathcal{G}, \mathcal{F})$\;
            $s_{\mathrm{evo}} \gets f(\pi_{\mathrm{evo}})$\;
            $v_{\mathrm{evo}} \gets (\pi_{\mathrm{evo}}, s_{\mathrm{evo}})$\;
            $\mathcal{V} \gets \mathcal{V} \cup \{v_{\mathrm{evo}}\}$\;
            $\mathcal{E} \gets \mathcal{E} \cup \{(v_{\mathrm{parent}}, v_{\mathrm{evo}})\}$\;
        }
        \Else{
            $\mathcal{V}_{\mathrm{top}} \gets \textsc{TopK}(\mathcal{G})$\;
            $\pi_{\mathrm{fused}} \gets \textsc{Fuse}(\mathcal{V}_{\mathrm{top}})$\;
            $s_{\mathrm{fused}} \gets f(\pi_{\mathrm{fused}})$\;
            $v_{\mathrm{fused}} \gets (\pi_{\mathrm{fused}}, s_{\mathrm{fused}})$\;
            $\mathcal{V} \gets \mathcal{V} \cup \{v_{\mathrm{fused}}\}$\;
        }
        $\mathcal{G} \gets (\mathcal{V}, \mathcal{E})$\;
    }
}

$(\pi^\star, s^\star) \gets \arg\max_{(\pi, s) \in \mathcal{V}} s$\;
\Return $\pi^\star$\;

\caption{\textbf{Monte Carlo Graph Search for training pipelines.} Each node stores a complete pipeline $\pi=(D,H,S)$ and its realized score $s=f(\pi)$. The agent iteratively expands promising leaves, evaluates proposed pipelines by actual train-and-test execution, backpropagates score statistics through the graph, and invokes evolution or cross-branch fusion when progress stalls.}
\label{alg:mcgs}
\end{algorithm}
\end{minipage}
\end{center}

\noindent\textbf{Notation.}
A training pipeline is $\pi=(D,H,S)$, where $D$ is the dataset specification, $H$ is the hyperparameter configuration, and $S$ is the learning strategy.
Each node $v=(\pi,s)$ stores a complete evaluated pipeline together with its realized validation-set score $s=f(\pi)$.
The search graph is $\mathcal{G}=(\mathcal{V},\mathcal{E})$, where $\mathcal{V}$ is the set of evaluated nodes and $\mathcal{E}$ records parent--child proposal relationships.
$\textsc{Leaf}(\mathcal{G})$ denotes the set of expandable leaf nodes, $\textsc{UCT}(v,\mathcal{G})$ is the selection score from Eq.~\ref{eq:uct}, $\textsc{Backprop}$ updates ancestor statistics after evaluation, $\textsc{TopK}(\mathcal{G})$ returns the top-scoring nodes, and $\textsc{Fuse}$ combines the pipelines contained in those nodes.

\textbf{Stagnation recovery.} When a branch stops improving (no score increase over several consecutive expansions), the agent triggers either \emph{evolution} (trajectory-aware mutation within the branch) or \emph{fusion} (cross-branch synthesis of the top-$K$ nodes).

\section{AdaptFT-Bench Details}
\label{app:adaptft-details}

\subsection{Construction Methodology}

\subsubsection{Scenario Selection}

AdaptFT-Bench has two components. The primary benchmark consists of seven benchmark-derived synthetic inference scenarios built from public datasets using our perturbation pipeline (Section~\ref{app:noise}). In addition, we report two production-style deployment case studies from Pioneer Agent deployments (CLINC150 and CoNLL-2003) to show the same loop on production traces. Across both components, scenarios span:

\begin{itemize}
    \item \textbf{Task types}: Binary classification, multi-class classification, NER, text generation
    \item \textbf{Model maturity}: Base models with many failures, fine-tuned models with residual failures
    \item \textbf{Model families}: GLiNER2 (encoder) and Qwen/Llama (decoder)
    \item \textbf{Domains}: SMS spam detection (classification), intent classification (multi-class), newswire NER, math reasoning (generation)
\end{itemize}

\subsubsection{Synthetic Inference Log Construction}

In addition to real production traces, we construct synthetic inference logs from benchmark tasks to study the same adaptation loop in a controlled and reproducible setting. Starting from benchmark examples, we generate perturbed user-style variants of the original inputs, run the target model on both the original and perturbed inputs, and store the resulting prompt-response pairs as inference traces. The perturbations span the main classes of production noise implemented in our pipeline, including linguistic errors, structural corruption, adversarial rewrites, off-task inputs, and repetitive or near-duplicate variants. The purpose of this construction is to provide realistic, diverse failure traces when real production logs are unavailable, while preserving the same downstream analysis, curriculum construction, and retraining procedure used for deployed systems.

\subsubsection{Failure Set Construction}

Failure sets come from two sources. For the seven synthetic scenarios, we perturb benchmark inputs, run the target model, and mark failures by comparing model outputs against the benchmark ground truth. For the two real case studies, we use judged production traces with LLM-as-judge verdicts and human overrides when available. In both settings, each failure set is constructed to satisfy the same criteria:
\begin{enumerate}
    \item Confirmed incorrect behavior with an associated gold or corrected output
    \item Diversity across at least three distinct failure categories
    \item Sufficient metadata to re-evaluate after training (benchmark ground truth for synthetic scenarios; judge criteria and prompt configuration for real case studies)
\end{enumerate}

\subsubsection{Regression Set Construction}

For each scenario:
\begin{enumerate}
    \item Sample passing inferences (\texttt{llmaj\_verdict = 'pass'} or \texttt{human\_verdict = 'correct'})
    \item Stratified by label/entity type to cover the model's existing strengths
    \item Size: 30--100\% of the failure set (ensures regression detection is sensitive while remaining practical for large failure sets)
\end{enumerate}

\subsubsection{Scenario Difficulty Tiers}

Table~\ref{tab:scenario-difficulty} defines the difficulty tiers used to categorize benchmark scenarios based on failure scale, diversity, and model maturity.

\begin{table}[htbp]
\centering
\begin{tabularx}{\linewidth}{
>{\raggedright\arraybackslash}l
>{\raggedright\arraybackslash}X
>{\raggedright\arraybackslash}l
>{\raggedright\arraybackslash}l
>{\raggedright\arraybackslash}l
}
\toprule
Tier & Description & Failure count & Failure diversity & Model state \\
\midrule
Easy & Few failures, clear patterns & 10--50 & 1--2 clusters & Fine-tuned \\
Medium & Many failures, systematic patterns & 50--500 & 3--5 clusters & Fine-tuned or base \\
Hard & Many failures, adversarial + non-fixable & 100--500+ & 5+ clusters, incl. non-fixable & Base model \\
\bottomrule
\end{tabularx}
\caption{\textbf{Scenario Difficulty Tiers in AdaptFT-Bench.}}
\label{tab:scenario-difficulty}
\end{table}

\subsubsection{Anonymization and Reproducibility}

\begin{itemize}
    \item All personally identifiable information (PII) and user-identifiable information is stripped from real inference logs
    \item The seven synthetic scenarios are reproducible from public benchmarks plus the released perturbation pipeline
    \item Base model weights are referenced by HuggingFace model ID (publicly available)
    \item Judge configurations (model, criteria, prompt template) are documented for the real case studies
    \item In this paper, CLINC150 and CoNLL-2003 are reported as anonymized case studies rather than released benchmark snapshots
\end{itemize}

\subsection{Scenario-Level Details}

Table~\ref{tab:adaptft-scenario-details} summarizes the scenario-level information needed to interpret the benchmark results in this paper. We do not reproduce the full raw failure taxonomies for every scenario here; instead, we provide the evaluation regime, metric, failure source, and the corresponding sanitized artifact when one is included in the representative bundle.

\begin{table}[htbp]
\centering
\small
\begin{tabularx}{\linewidth}{@{}l >{\raggedright\arraybackslash}X l >{\raggedright\arraybackslash}X >{\raggedright\arraybackslash}X >{\raggedright\arraybackslash}X@{}}
\toprule
Scenario & Regime & Base model & Metric & Failure source & Sanitized artifact \\
\midrule
CLINC150 & Production-style case study & GLiNER2-base & Exact-match pass rate on failing slice & 453 judged production failures + 198 passing regression examples & \texttt{clinc150\_trace\_and\_report\_summary.md} \\
CoNLL-2003 & Production-style case study & GLiNER2-base / large & Entity~F1 + sample pass rate & 2,740 judged NER failures from production-style deployment logs & \texttt{conll2003\_ner\_trace\_and\_report\_summary.md} \\
GSM8K (Qwen3-8B) & Synthetic 3-stage scenario & Qwen3-8B & Exact match on final \texttt{\#\#\#\#} number & Benchmark-derived noisy inference logs & none in representative bundle \\
GSM8K (Llama 3.2-3B) & Synthetic 3-stage scenario & Llama 3.2-3B & Exact match on final \texttt{\#\#\#\#} number & Benchmark-derived noisy inference logs & \texttt{gsm8k\_llama\_trace\_summary.md} \\
ARC-Challenge & Synthetic 3-stage scenario & Llama 3.2-3B & Exact match on answer letter & Benchmark-derived noisy inference logs & \texttt{arc\_challenge\_trace\_summary.md} \\
TriviaQA & Synthetic 3-stage scenario & Llama 3.2-3B & Case-insensitive answer-alias match & Benchmark-derived noisy inference logs & \texttt{triviaqa\_trace\_summary.md} \\
HumanEval & Synthetic 3-stage scenario & Qwen3-8B & pass@1 & Benchmark-derived noisy inference logs & none in representative bundle \\
XSum & Synthetic 3-stage scenario & Qwen3-8B & ROUGE-2 & Benchmark-derived noisy inference logs & none in representative bundle \\
SAMSum & Synthetic 3-stage scenario & Qwen3-8B & ROUGE-2 & Benchmark-derived noisy inference logs & none in representative bundle \\
\bottomrule
\end{tabularx}
\caption{Scenario-level details for AdaptFT-Bench as used in this paper. The seven synthetic scenarios all use the shared stage-based protocol from Section~\ref{sec:adaptft}: three stages with rising poison rates, 70/30 train-test splits per stage, and a held-out evaluation set formed by the union of stage test splits. CLINC150 and CoNLL-2003 are real judged deployment case studies reported separately from the synthetic inference benchmark.}
\label{tab:adaptft-scenario-details}
\end{table}

The derived artifact bundle intentionally covers only a representative subset of scenarios. We chose one encoder cold-start report (SMS Spam), three cold-start trace summaries (ARC-Challenge, GSM8K/Llama, TriviaQA), and two production-improvement case studies (CLINC150 and CoNLL-2003) because together they span the distinct trajectory motifs discussed in the paper: broad initial exploration, failure-taxonomy construction, curriculum synthesis, rollback, threshold search, and model-capacity escalation.

\subsection{Perturbation Pipeline}
\label{app:noise}

To evaluate production mode, we simulate realistic production traffic by injecting noise into benchmark inputs. The pipeline partitions each benchmark's training set into a clean training pool (used to produce the clean-trained checkpoint) and an inference pool (the remainder). Each sample in the inference pool is independently perturbed: every perturbation type rolls against its configured rate, so multiple noise types can co-occur on a single input. Temporal realism (log-normal latency jitter, batch effects, timeout/retry entries at 1.5\%) is applied to the resulting log sequence.

Table~\ref{tab:noise-types} lists all perturbation types and their default rates.

\begin{table}[htbp]
\centering
\caption{Noise injection rates for synthetic inference log generation. Each perturbation is applied independently per sample.}
\label{tab:noise-types}
\begin{tabular}{@{}lll@{}}
\toprule
Category & Perturbation Type & Default Rate \\
\midrule
\multirow{8}{*}{Linguistic}
 & Typos (character transposition/insertion/deletion) & 10\% \\
 & Phonetic misspellings & 7\% \\
 & Syntactic errors (broken structure, fragments) & 6\% \\
 & Grammatical errors (tense, articles, agreement) & 5\% \\
 & Punctuation abuse (missing, excessive) & 5\% \\
 & Casing errors (all-lower, all-caps, random) & 4\% \\
 & Abbreviations and slang & 3\% \\
 & Code-switching (partial foreign language) & 2\% \\
\midrule
\multirow{11}{*}{Structural}
 & Preamble injection (conversational opener) & 7\% \\
 & Instruction-following attempts (format requests) & 4\% \\
 & Truncated inputs (mid-sentence cutoff) & 3\% \\
 & HTML/markup remnants & 2\% \\
 & Missing modality reference & 2\% \\
 & Intra-input duplication & 1.5\% \\
 & Very long inputs ($>2\times$ median length) & 1.5\% \\
 & OCR artifacts (l/1, O/0 confusion) & 1.5\% \\
 & Empty or near-empty inputs & 0.7\% \\
 & UTF-8 anomalies (smart quotes, BOM, null bytes) & 0.7\% \\
 & Metadata leakage (benchmark provenance) & 0.7\% \\
\midrule
\multirow{6}{*}{Adversarial}
 & Label-flipping adversarials & 3\% \\
 & Decision-boundary probes & 3\% \\
 & False premise questions & 3\% \\
 & Prompt injection (in-content instructions) & 2\% \\
 & Cross-benchmark contamination (wrong endpoint) & 1.5\% \\
 & Jailbreak attempts & 1\% \\
\midrule
\multirow{6}{*}{Off-task}
 & Off-domain queries & 3\% \\
 & Meta-questions (``What model are you?'') & 2\% \\
 & Context-assumption queries (assumes state) & 1.5\% \\
 & Verbatim benchmark recall & 1\% \\
 & Gibberish and keyboard mashing & 0.7\% \\
 & Retry storms (3--8 rapid resubmissions) & 0.7\% \\
\midrule
\multirow{2}{*}{Repetition}
 & Near-duplicates (Jaccard $\geq$ 0.8) & 5\% \\
 & Exact duplicates & 3\% \\
\bottomrule
\end{tabular}
\end{table}

\section{Cold-Start Experimental Details}
\label{app:cold-start-trajectories}
\label{sec:sms-spam}

Table~\ref{tab:iteration-progression} summarizes performance at iterations 1, 3, and best across all cold-start benchmarks. On every task, the first training run captures only 40--70\% of the achievable improvement; the remaining gains come from the agent's ability to diagnose failures, adjust training strategy (e.g., switching to chain-of-thought supervision or tuning epoch count), and iterate.

\begin{table}[htbp]
\centering
\begin{tabular}{@{}llllll@{}}
\toprule
Benchmark & Model & Iter 1 & Iter 3 & Best & Iterations \\
\midrule
ARC-Challenge & Llama 3B & 48.1\% & 68.6\% & \textbf{72.6\%} & 11 \\
GSM8K & Llama 3B & 11.8\% & $\sim$35\% & \textbf{43.7\%} & 10 \\
HumanEval & Qwen 8B & 72.0\% & 85.4\% & \textbf{92.7\%} & 4 \\
XSum & Qwen 8B & R2: 13.0 & R2: 16.4 & \textbf{R2: 17.9} & 13 \\
SAMSum & Qwen 8B & R2: 21.8 & R2: 24.1 & \textbf{R2: 25.4} & 11 \\
TriviaQA & Llama 3B & 0.0\% & 44.0\% & \textbf{48.6\%} & 9 \\
\bottomrule
\end{tabular}
\caption{Cold-start performance progression across iterations. The number of iterations varies by task complexity, from 4 (HumanEval) to 13 (XSum).}
\label{tab:iteration-progression}
\end{table}

The per-benchmark narratives below detail the specific configuration changes at each iteration.

\paragraph{ARC-Challenge (Llama 3.2-3B).}
The base model cannot answer multiple-choice questions---it generates new questions instead. Starting from 5.3\%, the agent progressed through format learning (Iter~1: 48.1\%), data scaling (Iter~2: 61.3\%), CoT with GPT-4.1 (Iter~3: 68.6\%), CoT with DeepSeek-R1 (Iter~5: 68.9\%), and R1 CoT + validation data (Iter~6: 72.6\%). Iterations 7--11 (more data, higher rank, lower LR) all regressed to 68.7--72.4\%.

\paragraph{GSM8K (Llama 3.2-3B).}
Starting from $\sim$8\% (5-shot), the agent scaled from 300 examples (Iter~1: 11.8\%) to 7,473 examples over 5 epochs (Iters~2--5: 30--40\%), then found 2 epochs with batch size 16 optimal (Iter~6: 43.7\%). Iterations 7--10 with upweighting, different LR, or rank=64 could not surpass Iter~6.

\paragraph{HumanEval (Qwen3-8B).}
Trained on MBPP (374 problems), evaluated on HumanEval (164 different problems). The main result is 92.7\% pass@1 from MBPP data alone. In a separate exploratory run using oracle distillation (training on the model's own correct outputs), the agent reached 100\% pass@1 with 173 examples, but adding GPT-4.1 solutions regressed performance (96.9\%$\to$94.5\%).

\paragraph{XSum (Qwen3-8B).}
ROUGE-2 improved from 6.9 to 17.9 (+160\%). The agent found that 1 epoch is optimal and training beyond 750 gradient steps degrades performance.

\paragraph{SAMSum (Qwen3-8B).}
The largest single improvement came from system prompt engineering: replacing ``summarize this dialogue'' with a constrained specification. Final: ROUGE-1 50.5 (+36\%), ROUGE-2 25.4 (+89\%).

\paragraph{SMS Spam (GLiNER2-base).}
The agent downloaded the UCI SMS Spam Collection (5,574 messages, 87\% ham / 13\% spam), split 80/20, and launched 4 parallel jobs. Best initial model: F1 0.9834 with 5 residual failures. Three surgical augmentation rounds (+26, +12, +17 examples) drove F1 to 0.9967 via precision-recall balancing.

\subsection{Best-Model Hyperparameters}
\label{app:hparams}

\begin{table}[htbp]
\centering
\caption{Best training pipeline per benchmark, as selected autonomously by the agent. ``--'' indicates the agent used the platform default or the value was not recorded in the run artifacts.}
\label{tab:best-hparams}
\begin{tabular}{@{}lllllll@{}}
\toprule
Benchmark & Model & Train Ex. & Epochs & Batch Size & LR & Notes \\
\midrule
ARC-Challenge & Llama 3B & 1,119 & 5 & -- & -- & R1 CoT + val data \\
GSM8K & Llama 3B & 7,473 & 2 & 16 & 2e-4 & \\
TriviaQA & Llama 3B & 3,000 & 4 & -- & 1e-4 & \\
HumanEval & Qwen 8B & 374 & 3 & -- & -- & Cross-benchmark (MBPP) \\
XSum & Qwen 8B & 3,000 & 1 & 4 & -- & \\
SAMSum & Qwen 8B & 500 & 8 & -- & -- & Constrained system prompt \\
ARC-Challenge & Qwen 8B & -- & 5 & -- & 1e-4 & CoT \\
SMS Spam & GLiNER2-base & 4,513 & -- & -- & -- & Full FT + 55 augmented \\
\bottomrule
\end{tabular}
\end{table}

\section{Infrastructure and Reproducibility}
\label{app:infrastructure}

\begin{table}[htbp]
\centering
\begin{tabularx}{\linewidth}{@{}l >{\raggedright\arraybackslash}X@{}}
\toprule
Component & Configuration \\
\midrule
Agent LLM & Claude Sonnet 4.6, 1M context, 32K extended thinking budget \\
Agent framework & LangGraph + LangChain (Python) \\
Sandbox & Modal containers, 16GB memory, 24h timeout, 1 CPU \\
Training backend & Tinker SDK (LoRA fine-tuning, instant inference after training) \\
Training helpers & \texttt{tinker\_helpers.py} -- standalone wrappers for \texttt{train()}, \texttt{infer()}, \texttt{infer\_batch()} \\
Data generation LLMs & OpenAI GPT-4.1/4.1-mini, DeepSeek-R1/V3, Fireworks \\
Web search & Exa deep research API \\
Agent recursion limit & 1,500 turns (main), 1,000 turns (sub-agents) \\
Trace analysis (production mode) & PostgreSQL (Supabase) with per-user Row-Level Security (RLS) filtering \\
Sub-agent LLM & Claude Sonnet 4.6 (same as main agent) \\
\bottomrule
\end{tabularx}
\end{table}

\subsection{Tinker SDK Integration}

Cold-start mode uses the Tinker SDK directly for training and inference, bypassing any intermediary API. The agent's \texttt{tinker\_helpers.py} provides three core functions:

\begin{itemize}
    \item \textbf{\texttt{train(dataset\_path, base\_model, nr\_epochs, learning\_rate, batch\_size, lora\_rank)}} -- Executes the full LoRA training loop: tokenize $\rightarrow$ forward/backward $\rightarrow$ optimize $\rightarrow$ save weights. Returns a \texttt{weights\_ref} string for immediate inference.
    \item \textbf{\texttt{infer(prompt, weights\_ref, base\_model, ...)}} -- Creates a Tinker \texttt{SamplingClient} from the saved weights and generates text. No deployment step or external service required.
    \item \textbf{\texttt{infer\_batch(prompts, weights\_ref, ..., max\_workers=20)}} -- Parallel inference using \texttt{ThreadPoolExecutor} with 20 concurrent Tinker API calls.
\end{itemize}

Tokenization uses the model's own \texttt{apply\_chat\_template} for train/serve parity. Loss is computed only on assistant tokens via per-token loss weights. End-of-sequence (EOS) truncation is handled automatically for base models (Llama) that don't stop at end-of-turn tokens.

\subsection{Evaluation Code}

Each benchmark uses its own evaluation method, set up autonomously by the agent:

\begin{table}[htbp]
\centering
\begin{tabularx}{\linewidth}{@{}l >{\raggedright\arraybackslash}X >{\raggedright\arraybackslash}X@{}}
\toprule
Benchmark & Evaluation Method & Code Source \\
\midrule
ARC-Challenge & Exact match on answer letter (A/B/C/D) & Agent-written extraction + string match \\
GSM8K & Exact match on final number after \texttt{\#\#\#\#} & Agent-written regex extraction \\
TriviaQA & Case-insensitive match against answer aliases & Agent-written alias matching \\
HumanEval & pass@1 -- code execution against unit tests & Official \texttt{human\_eval} package \\
XSum & ROUGE-1, ROUGE-2, ROUGE-L & \texttt{rouge-score} Python package \\
SAMSum & ROUGE-1, ROUGE-2, ROUGE-L & \texttt{rouge-score} Python package \\
ARC-Challenge (Qwen) & Exact match on answer letter & Agent-written extraction \\
\bottomrule
\end{tabularx}
\end{table}
The agent downloads, installs, and configures evaluation code autonomously. For programmatic benchmarks (ARC, GSM8K, TriviaQA), the agent writes extraction functions. For standard metrics (ROUGE, pass@1), it installs and uses existing packages.

\subsection{Released Artifacts}
\label{app:artifacts}

We do not include the raw \texttt{data-curation.md} logs, raw JSON trace exports, or full LangSmith traces in the paper appendix. Those internal artifacts contain tool-level execution details, code, and implementation-specific prompts that are not needed to understand the scientific claims. Instead, we include a representative bundle of \emph{derived and sanitized} artifacts spanning both agent modes and the main task families in the paper: encoder classification, production repair, NER, reasoning, math, and knowledge QA.

\begin{table}[htbp]
\centering
\begin{tabularx}{\linewidth}{
>{\raggedright\arraybackslash}p{2.4cm}
>{\raggedright\arraybackslash}p{2.2cm}
>{\raggedright\arraybackslash}p{2.8cm}
>{\raggedright\arraybackslash}X
}
\toprule
Artifact family & Agent mode & Benchmarks & What it preserves \\
\midrule
Final reports (PDF) & Cold-start & SMS Spam & Domain research, dataset setup, 4-iteration augmentation loop, final F1 0.9967 \\
Final reports (PDF) & Production & CLINC150 & 5-cluster failure taxonomy, 1,254-example curriculum, Full FT vs.\ LoRA comparison, rollback decision \\
Final reports (PDF) & Production & CoNLL-2003 & FP-dominated diagnosis, 6-iteration repair loop, threshold sweep, capacity scale-up to GLiNER2-large \\
Derived trajectory summaries & Cold-start & ARC, GSM8K, TriviaQA & Search trajectories, iteration-by-iteration configuration changes, evaluation outputs, and intermediate decisions distilled from internal run exports \\
Derived trajectory summaries & Production & CLINC150, CoNLL-2003 & SQL diagnostics, delegated taxonomy analysis, dataset revisions, training/evaluation history, and stopping criteria distilled from internal traces \\
\bottomrule
\end{tabularx}
\end{table}

The current representative bundle includes:
\begin{itemize}
    \item \texttt{derived\_run\_artifacts\_md/README.md}
    \item \texttt{derived\_run\_artifacts\_md/sms\_spam\_report\_summary.md}
    \item \texttt{derived\_run\_artifacts\_md/clinc150\_trace\_and\_report\_summary.md}
    \item \texttt{derived\_run\_artifacts\_md/conll2003\_ner\_trace\_and\_report\_summary.md}
    \item \texttt{derived\_run\_artifacts\_md/arc\_challenge\_trace\_summary.md}
    \item \texttt{derived\_run\_artifacts\_md/gsm8k\_llama\_trace\_summary.md}
    \item \texttt{derived\_run\_artifacts\_md/triviaqa\_trace\_summary.md}
\end{itemize}

These artifacts are sufficient to reconstruct the main trajectory motifs discussed in Figure~\ref{fig:trajectory-motifs}: broad initial exploration, failure-taxonomy construction, targeted curriculum synthesis, rollback after regression, threshold search, and branch fusion. The raw internal traces were used only to derive these summaries and are not included verbatim in the paper or public artifact bundle.




\end{appendices}
\end{document}